\documentclass[acmtog,nonacm]{acmart}
\acmSubmissionID{191}
\usepackage{multirow}
\usepackage{booktabs} %
\usepackage{natbib}
\usepackage{graphicx}
\usepackage{float}
\usepackage{enumitem}
\usepackage{bbding}
\usepackage{soul}
\usepackage[ruled]{algorithm2e} %

\SetAlFnt{\small}
\SetAlCapFnt{\small}
\SetAlCapNameFnt{\small}
\SetAlCapHSkip{0pt}

\acmJournal{TOG}

\setcopyright{rightsretained}
\acmJournal{TOG}
\acmDOI{10.1145/3550454.3555445}

\begin{document}
\title{
Video-driven Neural Physically-based Facial Asset for Production
}

\author{Longwen Zhang}\authornote{Equal contributions.}
\orcid{0000-0001-8508-3359}
\affiliation{%
 \institution{ShanghaiTech University and Deemos Technology Co., Ltd.}
 \city{Shanghai}
 \country{China}}
\email{zhanglw2@shanghaitech.edu.cn}

\author{Chuxiao Zeng}\authornotemark[1]
\affiliation{%
 \institution{ShanghaiTech University and Deemos Technology Co., Ltd.}
 \city{Shanghai}
 \country{China}}
\email{zengchx@shanghaitech.edu.cn}

\author{Qixuan Zhang}\authornotemark[1]
\affiliation{%
 \institution{ShanghaiTech University and Deemos Technology Co., Ltd.}
 \city{Shanghai}
 \country{China}}
\email{zhangqx1@shanghaitech.edu.cn}

\author{Hongyang Lin}
\affiliation{%
 \institution{ShanghaiTech University and Deemos Technology Co., Ltd.}
 \city{Shanghai}
 \country{China}}
\email{linhy@shanghaitech.edu.cn}

\author{Ruixiang Cao}
\affiliation{%
 \institution{ShanghaiTech University and Deemos Technology Co., Ltd.}
 \city{Shanghai}
 \country{China}}
\email{caorx@shanghaitech.edu.cn}

\author{Wei Yang}
\affiliation{%
 \institution{Huazhong University of Science and Technology}
 \city{Wuhan}
 \country{China}}
\email{weiyangcs@hust.edu.cn}

\author{Lan Xu}\authornote{Corresponding author.}
\affiliation{%
 \institution{ShanghaiTech University and Shanghai Engineering Research Center of Intelligent Vision and Imaging}
 \city{Shanghai}
 \country{China}}
\email{xulan1@shanghaitech.edu.cn}

\author{Jingyi Yu}
\affiliation{%
 \institution{ShanghaiTech University and Shanghai Engineering Research Center of Intelligent Vision and Imaging}
 \city{Shanghai}
 \country{China}}
\email{yujingyi@shanghaitech.edu.cn}

\renewcommand\shortauthors{Zhang, L. et al}

\begin{abstract}
Production-level workflows for producing convincing 3D dynamic human faces have long relied on an assortment of labor-intensive tools for geometry and texture generation, motion capture and rigging, and expression synthesis. 
Recent neural approaches automate individual components but the corresponding latent representations cannot provide artists with explicit controls as in conventional tools. 
In this paper, we present a new learning-based, video-driven approach for generating dynamic facial geometries with high-quality physically-based assets. 
For data collection, we construct a hybrid multiview-photometric capture stage, coupling with ultra-fast video cameras to obtain raw 3D facial assets. 
We then set out to model the facial expression, geometry and physically-based textures using separate VAEs where we impose a global MLP based expression mapping across the latent spaces of respective networks, to preserve characteristics across respective attributes. %
We also model the delta information as wrinkle maps for the physically-based textures, achieving high-quality 4K dynamic textures. 
We demonstrate our approach in high-fidelity performer-specific facial capture and cross-identity facial motion retargeting. 
In addition, our multi-VAE-based neural asset, along with the fast adaptation schemes, can also be deployed to handle in-the-wild videos. 
Besides, we motivate the utility of our explicit facial disentangling strategy by providing various promising physically-based editing results with high realism. 
Comprehensive experiments show that our technique provides higher accuracy and visual fidelity than previous video-driven facial reconstruction and animation methods. 

\end{abstract}



\begin{teaserfigure}
    \setlength{\abovecaptionskip}{3pt}
    \centering
    \includegraphics[width=1\textwidth]{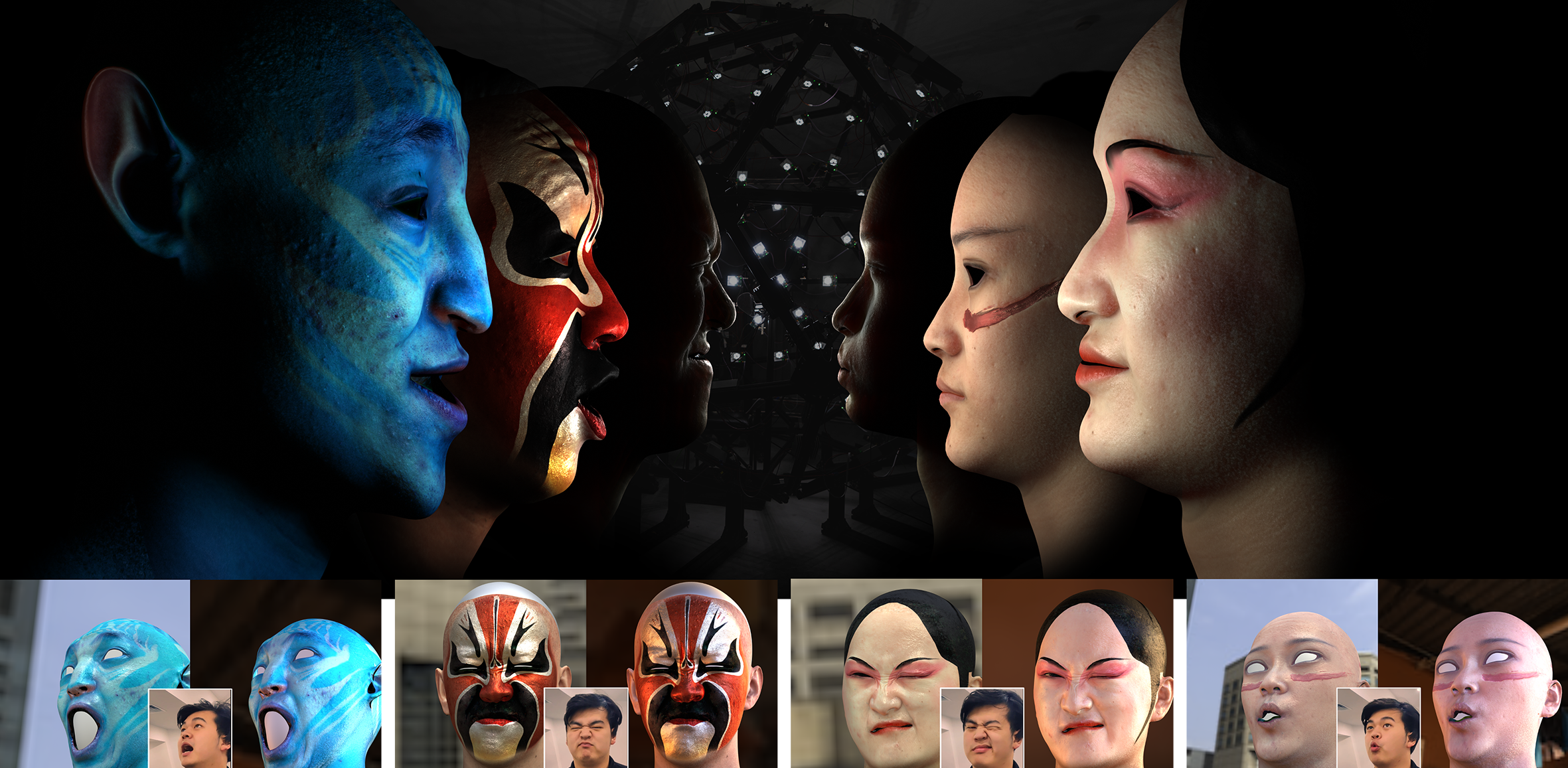}
    \caption{
{
Top: Our neural physically-based facial assets with stylized appearance. 
Bottom: cross-identity facial motion re-targeting driven by phone videos.
}
    }
    \label{fig:teaser}
\end{teaserfigure}

\maketitle

\section{Introduction}
High-quality and realistic digital avatars have been increasingly deployed in feature films, game productions, the entertainment industry, and most recently immersive experiences in the Metaverse. 
Despite tremendous advances in modeling 3D human body shapes ~\cite{SMPL:2015}, generating 3D dynamic facial assets at a production quality remains challenging: we humans are extremely sensitive to the perception of facial idiosyncrasies where even the slightest inconsistency can be easily detected. Successful solutions require not only reconstructing high-fidelity geometry, texture maps, normal maps, etc as in static facial asset generation but also reproducing realistic facial movement simulations indistinguishable from real videos. 

To avoid the uncanny valley, production-level workflows commonly employ {an assortment} of tools for building compelling and convincing dynamic facial assets. They range from scanning real 3D faces via expensive apparatus such as a Light Stage~\cite{debevec2000acquiring} or a multi-camera dome ~\cite{Joo_2017_TPAMI}, to conducting physically-based simulations to faithfully reproduce how light interacts with the soft tissues of the face in terms of appearance attributes (e.g., diffuse albedo, specular intensity, roughness, normal/displacement maps)~\cite{li2020learning}, and to capturing fine-grained facial motions~\cite{laine2017production} and transferring them onto a new performer with different physiognomy while performing a wide range of expressions~\cite{moser2021semi}. Once collected, the rich raw data at individual stages need to undergo time-consuming manual post-processing steps, 
where highly skilled artists need to carefully adjust both the physically-based textures and the facial expression models, e.g., following the Facial Action Coding System (FACS) ~\cite{ekman1978facial}. The standard pipeline is thus labor-intensive and expensive (in both cost and time) to be deployed to support a broader audience. To facilitate easier dynamic facial asset generations, there is an urgent demand to develop a more automatic workflow to at least reduce, if not fully eliminate, the manual labor in the production cycle. 
Recent deep learning techniques~\cite{li2020learning,li2020dynamic,lattas2021avatarme++} have provided a viable path to automated facial asset generation directly from the captured data with minimal human intervention.
In particular, the data-driven strategy employed by learning-based techniques is applicable to handling datasets of different scales, as many as thousands of subjects for producing a generic model or as few as a single subject with different expressions for portrait animation.
For the former, generic models~\cite{FLAME:SiggraphAsia2017,li2020dynamic,DECA:Siggraph2021,bao2020high,raj2021pixel} achieve high robustness despite environment lighting changes and variations in facial geometry and motion. %
Their generated assets are readily deployed to the existing facial animation production pipeline but they still rely on accurate blendshapes to achieve explicit control, {which often lead to loss of fine-grained details and cannot meet production-level quality.}
For the latter, recent performer-specific neural modeling tools can produce both geometry and appearance with fine details and conduct compelling end-to-end facial rendering ~\cite{Lombardi2019,VRface,Lombardi21,gafni2021dynamic,lombardi2018deep,cao2021real} at photo-realism. 
However, unlike generic models, neural approaches as an implicit representation cannot yet provide explicit controls over fine details in geometry, motion or physically-based textures, and therefore cannot be readily integrated into existing production pipelines for games or feature films, {which undermines the efficacy and efficiency of producing realistic and controllable facial assets. }

In this paper, we present a novel learning-based, video-driven approach for generating dynamic facial geometry along with high-quality physically-based textures including pore-level diffuse albedo, specular intensity and normal maps. %
Our approach supports both high-fidelity performer-specific facial capture and cross-identity facial motion retargeting. It is directly applicable for producing various physically-based effects such as geometry and material editing, facial feature (e.g., wrinkle) transfer, etc, as shown in Fig.~\ref{fig:teaser}.

\begin{figure*}[h]
    \centering
    \includegraphics[width=\linewidth]{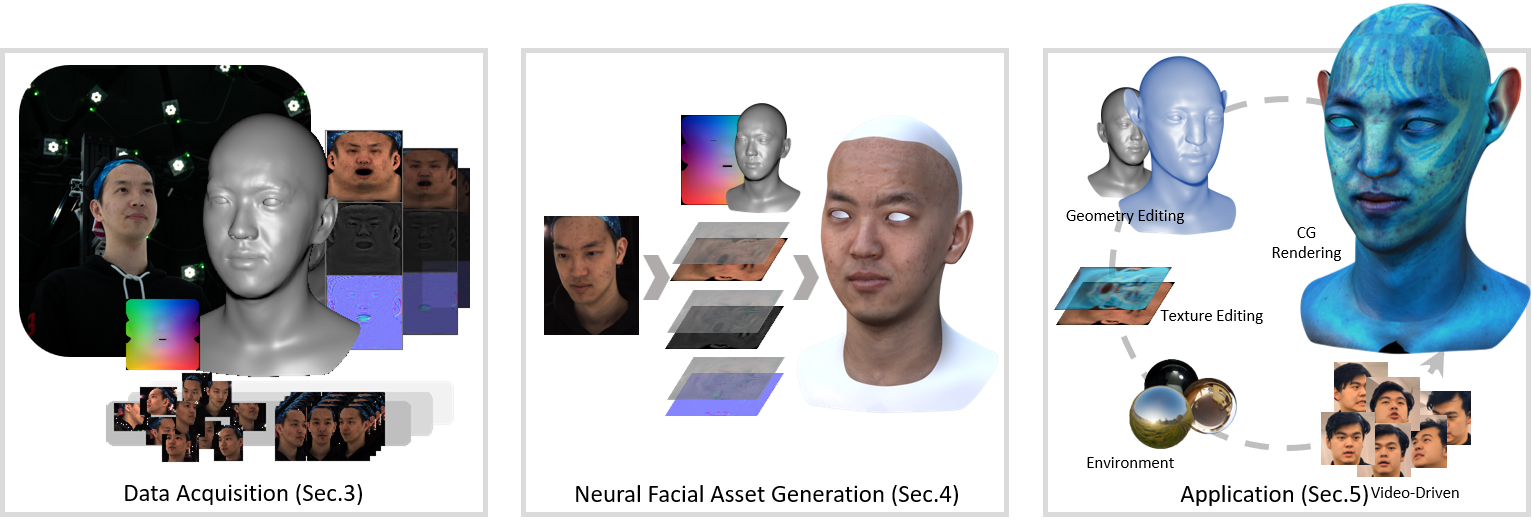}
    \caption{
{
\textbf{The neural physically-based facial asset generation pipeline.}
Our data acquisition system, i.e., the Fast Capture Stage (FaStage), extends the classical photometric LightStage~\cite{debevec2000acquiring} by combining multi-view cameras and photometric captures for recovering frame-by-frame detailed facial geometry and physically-based textures at a high frame rate. 
We then construct a neural representation of the facial asset using specifically designed multi-VAE architecture and training strategies, to simultaneously tackle geometry, textures, and expressions. The resulting neural asset, which is in neural representation, can be driven by in-the-wild face videos with matched facial motion and expression with production-level quality.}
    }
    \label{fig:overview}
\end{figure*}

We recognize that using videos, instead of separate shots, to produce physically-based facial assets imposes several challenges. On the data front, by far there are very few high-quality 3D facial video datasets publicly available. On processing, there lacks tailored neural modeling algorithms to handle the spatial-temporal facial data. {On editing, companion tools need to be developed.} %
We hence first construct a multi-view photometric fast capture stage or FaStage that extends previous solutions such as the USC Light Stage~\cite{debevec2000acquiring} to acquire facial geometry and textures under fast motion. FaStage employs a high-speed camera dome with 24 2K video cameras, a dense set of space-and-time-multiplexed illumination array, and three ultra-fast video cameras (Flex4K). The dome system is synchronized with the illumination pattern for obtaining proxy geometry whereas the Flex4K camera triplet can capture 38 gradient illumination patterns for normal and texture recovery at an ultra-high quality. 
We then apply Variational Autoencoders (VAE)~\cite{Kingma2014} to model the facial expression, geometry and physically-based textures of our assets via a unified neural representation. Different from prior work~\cite{lombardi2018deep,moser2021semi}, we introduce a multi-VAE framework to model the characteristics of individual facial attributes and maintain a well-structured latent space, so that the trained results can be applied to unseen new videos to support various video-driven facial applications.
{We further} employ a global Multi-Layer Perceptron (MLP) based expression mapping across the latent spaces of various VAEs to preserve characteristics across respective attributes.
Once trained, our neural facial asset can be directly applied to produce unseen facial animations, effectively enriching current scarce dynamic facial assets. Specifically, we present a fast adaptation scheme on the multi-VAE assets to enable cross-identity, fine-grained facial animations driven by in-the-wild video footage. 
Finally, unlike purely implicit representations, the disentangled facial attributes in our solution support direct geometry and material editing on well-established post-production tools, e.g., for producing stylized character animations. 
In the following sections, we provide details about our facial data acquisition and neural asset generation in Sec.~\ref{sec:data} and Sec.~\ref{sec:method}, respectively, followed by the performer-specific applications in Sec.~\ref{sec:application} (see Fig.~\ref{fig:overview} for the overview).

\section{Related Works}
Facial asset generation is a wide field, ranging from high-end productions that rely on sophisticated geometry and motion capture apparatus to low-grade-end consumer applications that use a single camera. For comprehensive reviews, we refer the readers to the recent surveys~\cite{morales2021survey,zollhofer2018state}. Our work aims to bridge the gap between the two by producing new facial assets from pre-captured ones driven by unseen, single-view videos, at production quality.

\vspace{-0.2cm}

\paragraph{Monocular Facial Reconstruction.} 
{Single-camera-based face animation methods generally adopt parametric face models with adaption for video observation based on landmark detection and head shape regression~\cite{blanz1999morphable, ekman1978facial}. Though effective and easy to manipulate, parametric models can not yet reach the highly realistic quality required for production.}
Blendshape is a prevalent parametric model where artists need to elaborately split or adorn the expressions for photorealistic re-creation and animation ~\cite{lewis2014practice}. 
It is also possible to register a template model to a real 3D scanned face of the performer under different expressions, e.g., by generating pre-defined blendshapes based on FaceWarehouse~\cite{Cao2014Facewarehosue} or 3D morphable models~\cite{egger20203dmm,smith2020morphable}.
\citet{weise2011realtime} fit the dynamic geometry of face to coarse 3D scans from RGBD cameras with pre-defined models whereas \citet{bao2020high} extend the approach for high-fidelity digital avatar generation from RGBD video inputs.
Various strategies have been proposed to improve the robustness and accuracy of single image facial asset generation ~\cite{cao20133d, paysan20093d, tuan2017regressing, guo2018cnn, richardson20163d, guo2020towards}. Alternative landmark-based reconstructions \cite{king2009dlib, zhang2014facial} are epitomized by the FLAME model ~\cite{FLAME:SiggraphAsia2017} that trains on a large dataset of spatial-time 4D scans. %
To further enhance reconstruction, \citet{cao2015real} set out to add high-fidelity facial details to low-resolution face tracking results. Yet one particular challenge is to reliably handle extreme skin deformation ~\cite{wu2016anatomically, chen2019photo, li2021topologically} while preserving realism.  DECA ~\cite{DECA:Siggraph2021} extends FLAME~\cite{FLAME:SiggraphAsia2017} by predicting performer-specific animatable details that can later be added back to the scanned facial assets of real performers. 
To animate facial assets, \citet{fyffe2014driving} drive high-quality scans with single-camera inputs whereas \citet{laine2017production} build a learning-based performer-specific framework for production. Various subsequent techniques ~\cite{chen2021high,ma2021pixel} also support fine-grained performer-specific facial capture. %
\citet{olszewski2016high} process multiple identities within limited viewpoints whose faces are covered by head-mounted displays. %
\citet{moser2021semi} apply image-to-image translation and extract a common representation for the input video and rendered CG sequence to predict the weights of the target character's PCA blendshapes.

\paragraph{Physically-based Texture Generation.} 
In production, physically-based textures such as diffuse albedo, normal maps, specular maps and displacement maps have long served as the key ingredients for creating photo-realistic digital avatars. 
\citet{ma2007rapid} pioneer using gradient illumination and diffuse-specular separation with polarizers to acquire high-fidelity physically-based textures. By changing the arrangement of linear polarizers, ~\citet{ghosh2011multiview} successfully adapt the previous method to a multiview setting.  To support dynamic expressions, \citet{fyffe2011comprehensive} use Phantom v640 high-speed cameras to achieve facial performance capture at 264 fps.  \citet{fyffe2015single} use a color polarized illumination setup to enable texture scanning in a single shot. \citet{legendre2018efficient} modify the camera setting to a mono camera to achieve more efficient and higher-resolution results. \citet{riviere2020single} manage to produce high-quality appearance maps from a single exposure. As for full-body performance capture, \citet{guo2019relightables} combine spherical gradient illumination with volumetric captures to produce compelling photorealism. With the advances in deep learning, possibilities of estimating the parameters of a predefined reflectance field from single image input ~\cite{saito2017photorealistic, huynh2018mesoscopic,yamaguchi2018high} have been demonstrated.

Passive facial capture has also made significant progress in recent years, largely attributed to the open-source multi-view reconstruction software ~\cite{beeler2010high}. By including skin parameters such as skin color and variations in hemoglobin concentration,
\citet{gotardo2018practical} manage to obtain more detailed facial features. However, in general, data produced by passive capture methods still cannot match the photometric ones due to calibration errors, camera resolutions, reconstruction errors, etc.
 
Emerging approaches have employed generative adversarial networks (GANs) to synthesize or infer textures of competent quality for industrial uses. \citet{li2020learning} introduce a framework to generate geometry and physically-based textures from identity and expression latent spaces, with super-resolution networks to produce textures that contain pore-level geometry.  \citet{lattas2021avatarme++} manage to infer renderable photorealistic 3D faces from a single image. 
Blendshapes can also be predicted from a single 3D scan of the neutral expression of the performer and then infer dynamic texture maps in a generative manner based on expression offsets ~\cite{li2020dynamic}.
Differently, we build dynamic textures using wrinkle maps, which allows us to maintain the authenticity of the high-resolution textures while avoiding the artifacts generated by super-resolution networks.

\begin{figure*}[t]
    \centering
    \includegraphics[width=\linewidth]{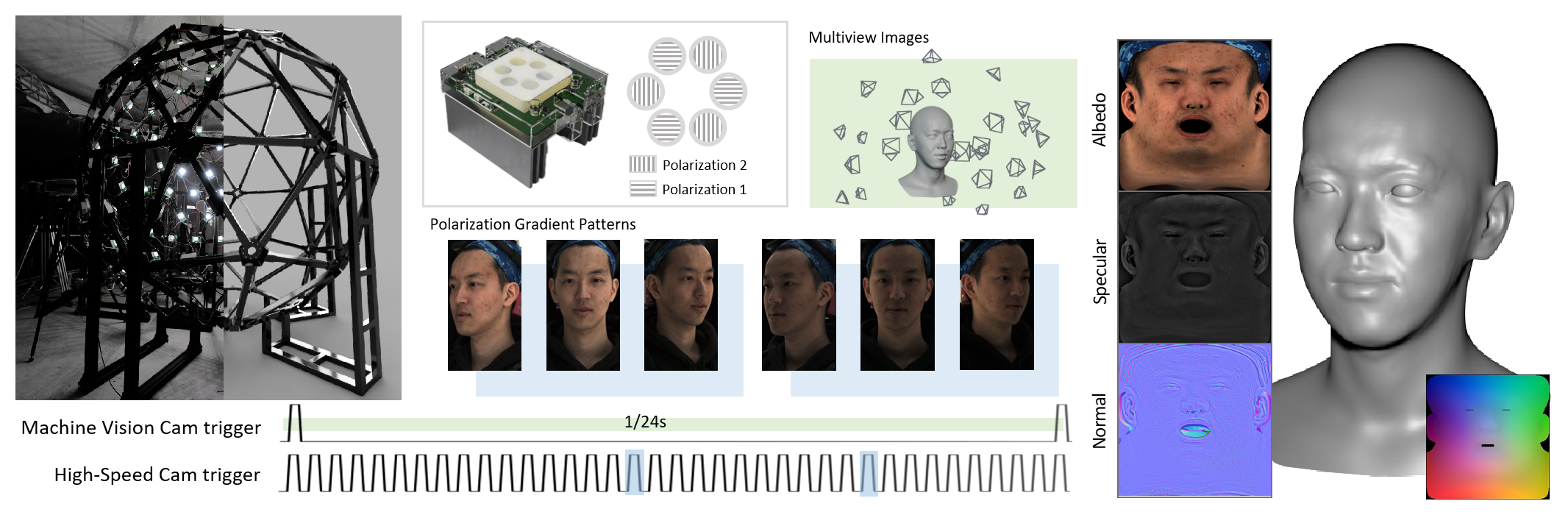}
    \caption{
\textbf{Illustration of our data acquisition pipeline.}
FaStage uses three high-speed cameras to capture polarization gradient patterns for dynamic physically-based texture acquisition. Multiple machine vision cameras are synchronized at a low frequency for dynamic facial geometry acquisition.
}
\label{fig:data_acquisition}
\end{figure*}

\vspace{-0.2cm}

\paragraph{Deep Face Models.} It has become increasingly accessible to replace classical geometry and texture-based models with deep models. A neural network can potentially extract a latent vector from the image input and later decode it into high-quality mesh and textures.
{\citet{chandran2020semantic} provide a parametric face model with semantic control over the identity and expression of the subject. \citet{lombardi2018deep} use a conditional variational autoencoder (CVAE) framework~\cite{Kingma2014} to encode geometry and view-dependent texture information. %
\citet{yoon2019self} utilize this model 
and make it capable of handling monocular input images under uncontrolled lighting environments. \citet{cao2021real} further estimate a gain map and a bias map to additionally improve rendering. \cite{cao2022authentic} propose Universal Prior Model (UPM) and produce photo-realistic performer-specific avatars for unseen identities. }

GAN-based methods ~\cite{gecer2019ganfit,li2020learning,lattas2021avatarme++} separate the shape and expression or using photorealistic differentiable-rendering-based training to enhance geometry details. %
Other identity-agnostic methods ~\cite{nirkin2019fsgan, abrevaya2020cross,burkov2020neural} aim to enhance the identity-independence by imposing constraints on the latent space. \citet{feng2018joint} introduce the UV-position map for 3D geometry representation to enable better regression of CNN.  \citet{bi2021deep} build a deep face model capable of producing OLAT {(one-light-at-a-time)} textures, making the predicted avatar relightable under novel lighting environments. In our work, we further improve the quality and editability of the deep face models by generating physically-based textures, owing to the high-quality dynamic facial assets we captured.

\section{Dynamic Facial Asset Acquisition}
\label{sec:data}

\subsection{The FaStage} \label{Sec:3.1}
To construct a performer-specific neural facial asset for product-level rendering, the most renowned system, the USC Light Stage~\cite{debevec2000acquiring}, exploits the time-multiplexed lighting to recover the facial reflectance field at unprecedented high quality. As shown in Fig.~\ref{fig:data_acquisition}, our FaStage adopts a similar semi-dome structure with evenly mounted space-and-time-multiplexed light units. Distinctively, in our FaStage, we replace the DSLR cameras in the USC Light Stage with 3 ultra-fast video cameras and 24 2K video cameras to capture the photos of the subject under 38 gradient illumination patterns and faithfully reconstruct the physically-based dynamic facial performance at video frame rate. 
We further provide the detailed design and a thorough discussion about our FaStage and captured assets with previous systems \cite{3drfe,yang2020facescape,Cosker2011,Cheng2018,yin20063d,yin20084d,yin2009bjut,Cao2014Facewarehosue,zhang2014bp4d,fyffe2014driving} in the supplementary material.

\subsection{Facial Asset Collection} \label{Sec:3.2}
Here we provide the details of collecting topology-consistent assets for typical facial movements at video frame rate, which is critical for exploiting deep networks in our neural facial asset training. Specifically, we capture a total of 5 minutes of facial performance for each performer at 24 fps and acquire the corresponding facial assets. We designed our expression span similar to previous works~\cite{ekman1978facial,laine2017production}, including rich expressions covering sufficient combinations of facial muscle motions {(please refer to the supplementary material)}.

For these representative performances, we reconstruct the high-quality dynamic facial assets including facial geometry, diffuse albedo, specular intensity, and normal map from the data captured by our FaStage. During capture, we send one specific control signal to turn all the lights on and collect the full-lit images from both the machine vision and high-speed cameras. Then, we perform the multi-view stereo technique on full-lit images to extract the facial geometry and fit a human face model~\cite{li2020learning} to obtain topology-aligned results similar to previous work~\cite{Cosker2011}. We also follow the work of ~\citet{li2020learning} to unwarp the facial geometry into UV maps and uniformly sample 256$\times$256 points where each pixel represents the location of the corresponding point on the mesh to generate geometry maps. To obtain physically-based textures, we apply 38 gradient illumination patterns for every 1/24 second and then use the 38 images from high-speed cameras under the corresponding patterns to recover diffuse albedo, specular intensity, and object-space normal map at 24 fps using the photometric stereo technique~\cite{ma2007rapid}. We adopt the standard fusion method~\cite{ghosh2011multiview} to merge the corresponding texture across different views in UV space and further transform the object-space normal maps into tangent space to reduce temporal noise.

\paragraph{Temporal Stabilization.}  
\begin{figure}[t]
    \centering
    \includegraphics[width=\linewidth]{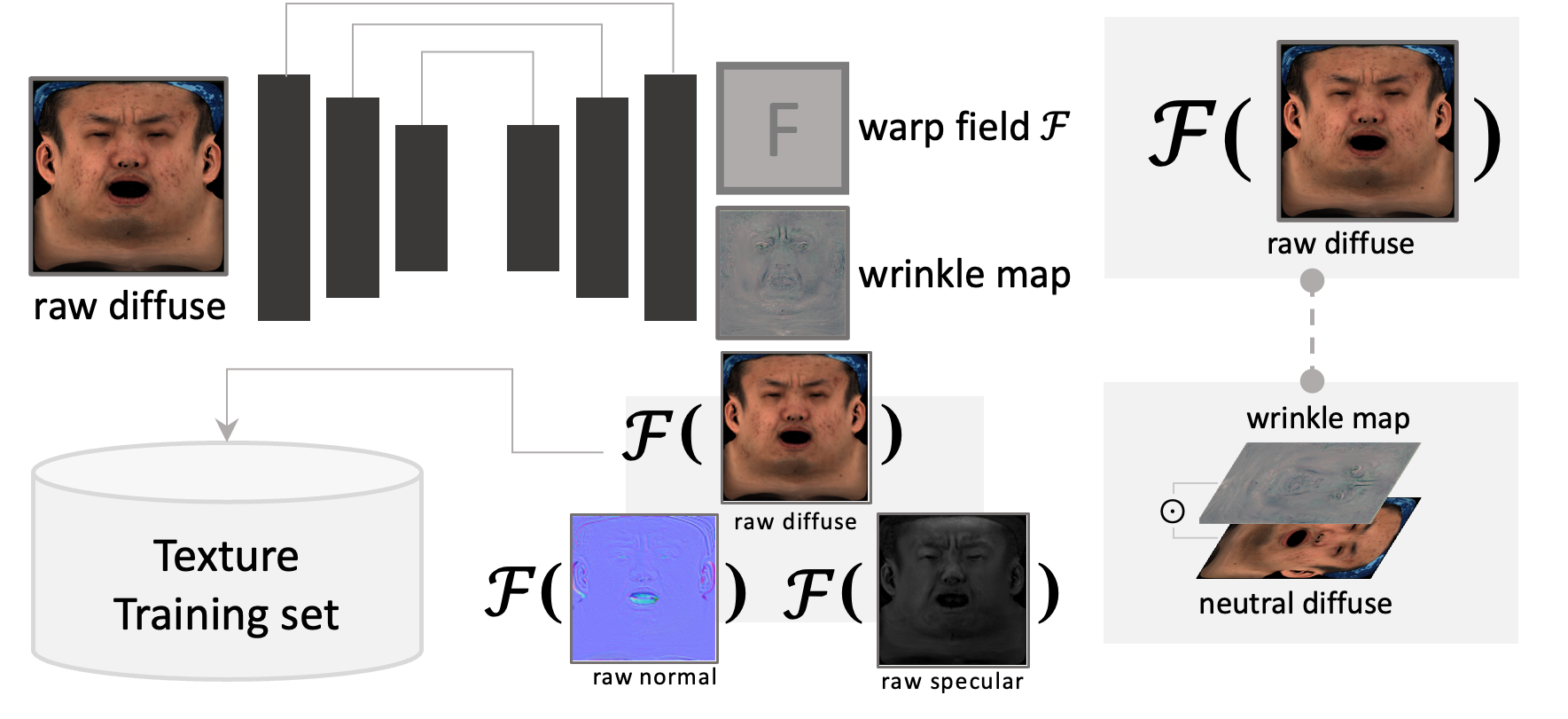}
    \caption{
\textbf{Pipeline of texture stabilization.}
We utilize an unsupervised method to predict the warp field with regard to the neutral textures for textures of each frame.
Stabilized textures, which are produced by applying the warp field to the textures of each frame, will be used as training data.
    }
    \label{fig:flow_net}
\end{figure}

\begin{figure*}[h]
    \centering
    \includegraphics[width=\linewidth]{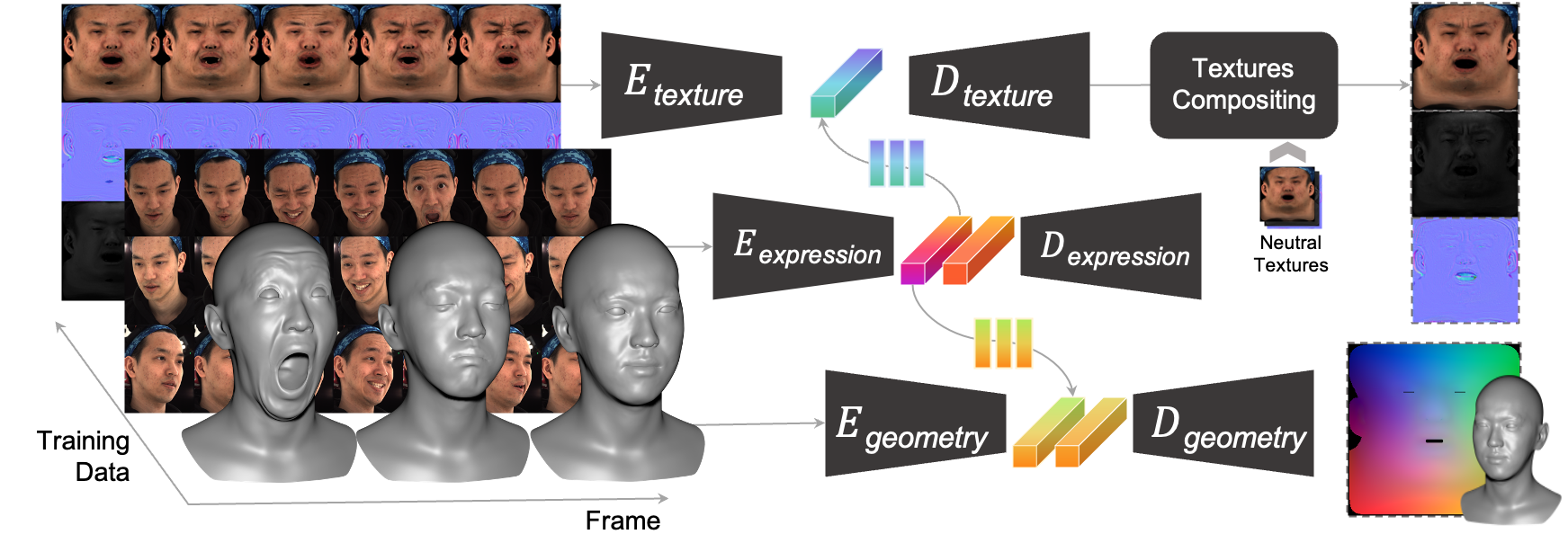}
    \caption{
\textbf{Neural physically-based facial asset.} For each frame, the training input consists of multi-view facial images captured in FaStage, as well as the coarse geometry and physically-based textures. Our approach trains three deep Variational Auto-Encoders to obtain three latent spaces separately on captured image, geometry and textures. Then two multilayer perceptrons are trained for latent space translation from expression to geometry and textures.}
    \label{fig:method}
\end{figure*}

We recognize the noticeable temporal jittery artifacts on the captured physically-based textures after the above unwrapping and multi-view merging process, especially for our 4D capture setting at video frame rate, leading to unreliable neural asset training. Thus, we introduce a novel self-supervised scheme to stabilize the obtained temporal textures, as illustrated in Fig.~\ref{fig:flow_net}. We adopt a U-Net structure to predict the temporal warp fields and wrinkle maps for the original textures in each frame with regard to the neutral one. Since the subtle appearance details in diffuse albedo are more suitable for temporal alignment, we only adopt the diffuse albedo for our stabilization scheme. To this end, given an input diffuse albedo $\mathbf{T}_t$ at $t$-th frame, our stabilization network $\mathbf{\Phi}$ predicts both a warp field $\mathcal{F}_t$ and a corresponding wrinkle map $\mathcal{W}_t$:
\begin{equation}
	\mathcal{F}_t, \mathcal{W}_t=\mathbf{\Phi}(\mathbf{T}_t).
\end{equation}
Here, $\mathcal{F}_t$ represents pixel-wise flow and we use $\mathcal{F}_t(\cdot)$ to represent the warping of a texture to the neutral frame where the performer makes no expressions. The wrinkle map $\mathcal{W}_t$ denotes the element-wise multiplication maps on top of the neutral texture in the UV domain to represent the stretch and squeeze of wrinkles from neutral to current expression caused by facial motions. It ranges from 0 to 1 to maintain the original brightness of captured diffuse albedo. Then, we formulate the following fidelity loss to ensure temporally aligned diffuse albedo after applying the warp field $\mathcal{F}_t$ to $\mathbf{T}_t$: 
\begin{equation}
	\mathcal{L}_{fid}=\|\mathcal{F}_t(\mathbf{T}_t)-\mathcal{W}_t\odot \mathbf{N} \|_1,
\end{equation}
where $\mathbf{N}$ is the neutral diffuse albedo and $\odot$ is element-wise multiplication.
We also adopt a temporal consistency term to ensure texture stability between consecutive frames to implicitly align two textures.
Given the network output $\mathcal{W}_t,\mathcal{W}_{t+1}$ from consecutive frames, the temporal term is formulated as:
\begin{equation}
	\mathcal{L}_{tem}=\|\mathcal{W}_t-\mathcal{W}_{t+1}\|_1.
\end{equation}
In addition, we apply a regularization term to restrict and regularize the estimated flow and its gradient:
\begin{equation}
	\mathcal{L}_{dis}=\|\mathcal{F}_t\|_2+\|\nabla \mathcal{F}_t\|_2.
\end{equation}
Our total loss for texture stabilization is formulated as follows:
\begin{equation}
	\mathcal{L}_{stab} = \mathcal{L}_{fid}+\mathcal{L}_{tem}+\mathcal{L}_{dis},
	\label{eq:loss_flow_net}
\end{equation}
where weights for balancing different losses are omitted for the simplicity of presentation.
Note that the diffuse albedo, specular intensity, and normal map share the same warp field, therefore we apply $\mathcal{F}$ to all the other texture attributes similar to the diffuse albedo.
For training our network $\Phi$, we use all the captured dynamic textures to optimize $\mathcal{L}_{stab}$. We downsample the input textures to 512$\times$512 to predict the corresponding warp fields and wrinkle maps. We implement our U-Net architecture and perform the training process according to Pytorch-UNet~\cite{Pytorch-UNet}.
Once trained, we apply the warp fields from $\Phi$ to process all the textures to obtain stabilized ones for the generation of neural facial assets in Sec.~\ref{sec:method}.

\section{Neural Facial Asset Generation}
\label{sec:method}

From captured dynamic facial assets of a specific performer, we intend to transform them into a drivable neural representation.
Traditional human facial performance capture relies on heavy manual labor to align geometry and textures, rigging, and expression deformation. 
This process can be potentially bypassed by a neural-based approach.
To this end, we build a learning framework with three networks for disentangling the expression, facial geometry and physically-based texture. 
The whole training pipeline is illustrated in Fig.~\ref{fig:method}.

\subsection{Expression and Viewpoint Disentanglement}
\label{sec:4.4}

\begin{figure}[t]
    \centering
    \includegraphics[width=\linewidth]{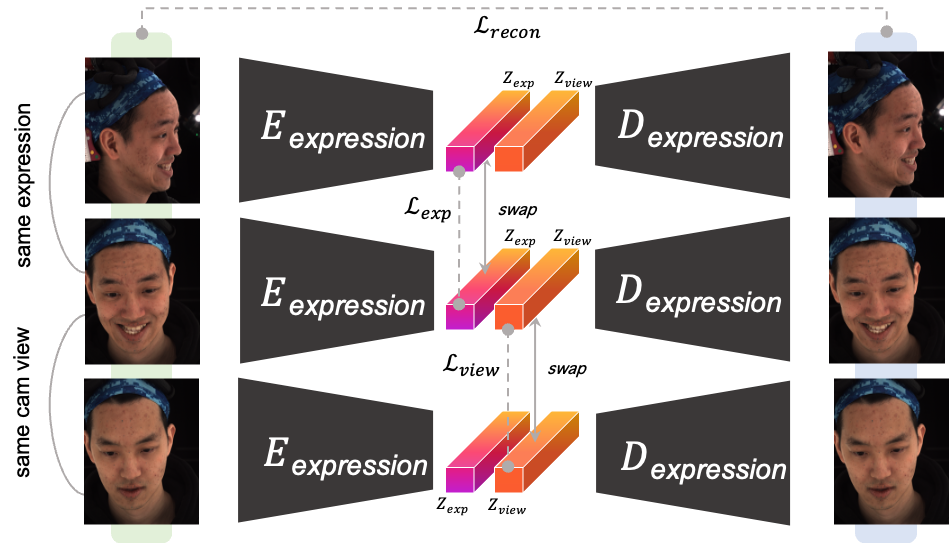}
    \caption{
\textbf{Architecture of expression network.}
{We constrain the encoder to disentangle expression and view.}
    }
    \label{fig:expression_network}
\end{figure}

We train a VAE to learn a global representation for expressions and viewpoints from training footage, as illustrated in Fig.~\ref{fig:expression_network}.
Given an input image ${I}$ from view $i$ at frame $t$, our expression encoder $\mathcal{E}_{e}$ encodes it to an expression latent code ${Z}_{e}$ and a viewpoint latent code ${Z}_{v}$,
while our expression decoder $\mathcal{D}_e$ predicts the reconstructed image ${\hat I}$ given these two latent codes.
To ensure latent space consistency between the encoder and decoder, we predict cycled latent codes ${\hat Z}_{e},{\hat Z}_{v}$, ($\hat{Z} = \mathcal{E} [\mathcal{D} (Z) ]$), from reconstructed image ${\hat I}$, and adopt the reconstruction and latent consistency loss:
\begin{equation}
\mathcal{L}_{rec}=\|{I}-{\hat I}\|_1, \, \, \mathcal{L}_{cyc}=
\|{Z}_{e}-{\hat Z}_{e}\|_2+
\|{Z}_{v}-{\hat Z}_{v}\|_2 .
\label{eq:Lrecon}
\end{equation}

\noindent Notice we omit view and frame index $i$ and $t$ in the equation for simplicity in presentation.

\vspace{3mm}\noindent{\bf Triplet supervision for expression encoding.}  \label{sec:trip_exp}
We designed a triplet training scheme which applies a latent swapping method ~\cite{huang2018multimodal,aberman2019learning, park2020swapping} that focuses on retrieving consistent expression and viewpoint information across views and frames. 
In each training iteration, we randomly pick two frames $s,t$, and two random views $i,j$. Notice all frames and views data are from the same performer.
Given the input images ${I}^{t,i}$, ${I}^{t,j}$, ${I}^{s,i}$, our encoder $\mathcal{E}_{e}$ predicts the performer's expressions and the camera's viewpoint latent codes for each of them, i.e., ${Z}^{t,i}$, ${Z}^{t,j}$, ${Z}^{s,i}$:
Notice the performer exhibits the same expression in ${I}^{s, i}$ and ${I}^{s, j}$, while in ${I}^{s, i}$ and ${I}^{t, i}$, the viewpoints are the same. 
Therefore, we minimize the distance between the expression latent code and the view latent code to achieve expression-viewpoint disentanglement:
\begin{equation}
\mathcal{L}_{exp}=\| {Z}^{t,i}_{e}-{Z}^{t,j}_{e}\|_2, \, \,
\mathcal{L}_{view}=\|{Z}^{t,i}_{v}-{Z}^{s,i}_{v}\|_2 .
\end{equation}
Besides constraining the encoder, we want the decoder to focus on recovering the expression despite the disturbances from viewpoint changes.
Therefore, the subsequent reconstruction is done in a cross frame and view manner:
\begin{equation}
\begin{aligned}
    {\tilde I}^{t,i}&=\mathcal{D}_{e}({Z}^{t,j}_{e}, {Z}^{s,i}_{v}),\\
    {\tilde I}^{t,j}&=\mathcal{D}_{e}({Z}^{t,i}_{e}, {Z}^{t,j}_{v}),\\
    {\tilde I}^{s,i}&=\mathcal{D}_{e}({Z}^{s,i}_{e}, {Z}^{t,i}_{v}) .
\end{aligned}
\end{equation}
The decoded result ${\tilde I}^{t,i}$ should be the same as the original image ${I}^{t,i}$, since the encoded expression and viewpoint information are supposed to be the same. 
The corresponding reconstruction error 
at this stage is formulated as:
\begin{equation}
\mathcal{L}_{cro}=
\|{I}^{t,i}-{\tilde I}^{t,i}\|_1+
\|{I}^{t,j}-{\tilde I}^{t,j}\|_1+
\|{I}^{s,i}-{\tilde I}^{s,i}\|_1.
\label{eq:loss_expression_cross}
\end{equation}
Then the total loss for our expression network can be formulated as follows:
\begin{equation}
\mathcal{L}_{E}=\mathcal{L}_{rec}+\mathcal{L}_{cyc}+\mathcal{L}_{exp}+\mathcal{L}_{view}+\mathcal{L}_{cro}.
\label{eq:loss_expression_net}
\end{equation}
This loss helps us to constraint $Z_e$ so that it only contains the expression features, and the viewpoint information is in $Z_v$. Notice we omit balancing weights for simplicity in presentation.

\subsection{Geometry Extraction}
\label{sec:4.5}

\begin{figure}[t]
    \centering
    \includegraphics[width=\linewidth]{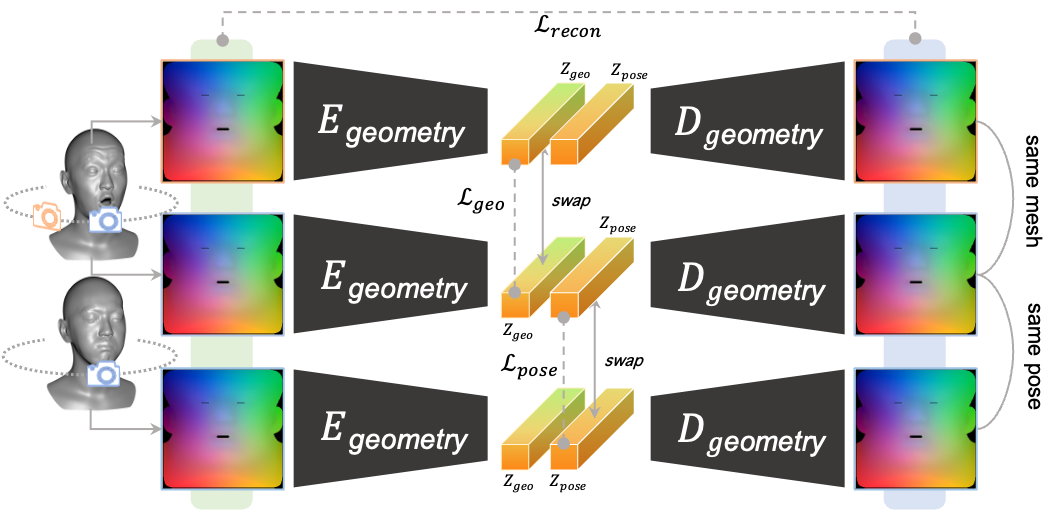}
    \caption{
\textbf{Architecture of geometry network.}
{We constrain the encoder to disentangle geometry and pose.}
    }
    \label{fig:geometry_network}
\end{figure}

Recall that we represent the facial geometry with a 2D-geometry map in UV space where each pixel represents the 3D coordinate of the corresponding vertex. Therefore, we can regard geometry inference as an image reconstruction task.
To extract better geometry features and eliminate the minor head pose errors, we adopt a similar network structure to the one developed for expression encoding. For improving robustness, we apply random rotations and translations to the scanned geometries for data augmentation.
As illustrated in Fig.~\ref{fig:geometry_network}, given an input geometry map ${G}$ from frame $t$ with augmentation operation $a$, our geometry encoder $\mathcal{E}_{g}$ encodes it to a geometry latent code ${Z}^{t,a}_{g}$ and a pose latent code $Z^{t,a}_{p}$,
while our geometry decoder $\mathcal{D}_{g}$ predicts the reconstructed image ${\hat G}^{t,a}$ given these two latent codes. 
Similar to the expression disentanglement, we randomly pick frames with $s,t$ and augmentation operations $a, b$ for training. Note that we choose $s,t$ from the same dynamic clip and make sure they are temporally close to ensure both geometries without augmentation have similar poses.
Specifically, when we fix the geometry latent code and change the pose latent code, the geometry maps generated by the decoder should exhibit similar geometry and only differ in rotation and translation.
Therefore, the decoding process itself is designed to be independent of the pose latent code, which could be formulated as:
\begin{equation}
    \mathcal{D}_{g}({Z}_{g}, {Z}_{p})=
    \mathbf{M}_p({Z}_{p}) \otimes \mathcal{D^\dagger}_{g}({Z}_{g}),
\end{equation}
where $\mathbf{M}_p$ is a multilayer perceptron (MLP) for transforming the pose latent code with a rotation and translation matrix, $\mathbf{D^\dagger}_{g}$ represents the network only for decoding the geometry latent code, and $\otimes$ represents matrix-vector multiplication. Notice we omit the frame index $t$ and the augmentation operation $a$ for simplicity.
In this way, our network is able to predict the performer's facial geometry free from the disturbance of head pose variance.

We use the same triplet supervision strategy as the expression network to disentangle the facial geometry and head pose, which has the following metrics:
\begin{equation}
\begin{aligned}
&\mathcal{L}_{rec}=\|{G}^{t,a}-{\hat G}^{t,a}\|_2,\\
&\mathcal{L}_{cyc}=
\|{Z}^{t,a}_{g}-{\hat Z}^{t,a}_{g}\|_2+
\|{Z}^{t,a}_{p}-{\hat Z}^{t,a}_{p}\|_2,\\
&\mathcal{L}_{geo}=\| {Z}^{t,a}_{g}-{Z}^{t,b}_{g}\|_2,\\
&\mathcal{L}_{pos}=\|{Z}^{t,a}_{p}-{Z}^{s,a}_{p}\|_2,\\
&\mathcal{L}_{cro}=
\|{G}^{t,a}-{\tilde G}^{t,a}\|_2+
\|{G}^{t,b}-{\tilde G}^{t,b}\|_2+
\|{G}^{s,a}-{\tilde G}^{s,a}\|_2 .
\end{aligned}
\end{equation}
Where $\mathcal{L}_{rec}$, $\mathcal{L}_{cyc}$ are reconstruction and cycled latent loss. $\mathcal{L}_{geo}$ and $\mathcal{L}_{pos}$ are the distances between latent codes. $\mathcal{L}_{cro}$ is the cross inference loss, similar to the definition in Eq.~\ref{eq:loss_expression_cross}. The total loss for the geometry network is formulated as follows:
\begin{equation}
\mathcal{L}_{G}=\mathcal{L}_{rec}+\mathcal{L}_{cyc}+\mathcal{L}_{geo}+\mathcal{L}_{pos}+\mathcal{L}_{cro}.
\label{eq:loss_geometry_net}
\end{equation}

\subsection{Learning Textures}
\label{sec:4.6}

\begin{figure}[t]
    \centering
    \includegraphics[width=\linewidth]{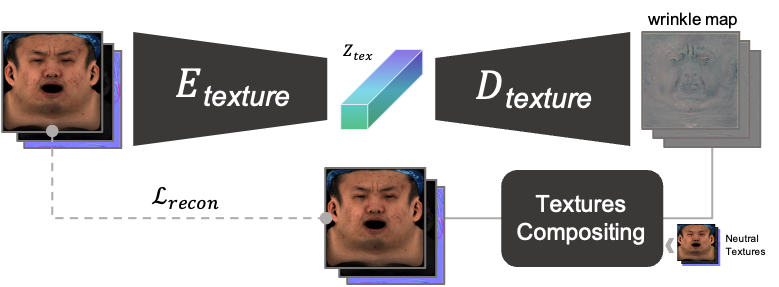}
\caption{
\textbf{Architecture of texture network.}
To enable high-resolution dynamic textures, our texture network outputs wrinkle maps instead of textures themselves.
}
    \label{fig:texture_network}
\end{figure}

The stabilized textures are view-independent and only contain expression information. Therefore, we build a similar VAE with only one latent space to learn the representations for the three types of textures (diffuse albedo, specular intensity, and normal maps) simultaneously. The network structure is illustrated in Fig.~\ref{fig:texture_network}. Notice the captured texture is at 4K resolution, yet the output resolution of the network is typically 512, which inevitably leads to a significant loss of details. To retain the ability of rendering and editing at 4K resolution, we enforce this network to predict wrinkle maps (we borrow this term from the CG industry
) instead of predicting the textures directly.
Given an input texture ${T}$ at frame $t$, our texture encoder $\mathcal{E}_{tex}$ encodes it to a texture latent code ${Z}_{tex}$, 
while our texture decoder $\mathcal{D}_{tex}$ predicts the corresponding wrinkle maps ${\hat W}$ from the texture latent code ${Z}_{tex}$:
\begin{equation}
    {\hat W}=\mathcal{D}_{tex}({Z}_{tex}), \, \, {\hat T} ={\hat W} \oplus \mathbf{N},
\end{equation}
where $\mathbf{N}$ is the neutral texture, ${\hat T}$ represents the reconstructed textures, and $\oplus$ represents the addition operation.
Similarly, we compute the texture reconstruction loss and cycled latent loss. 
During the training process, the losses are formulated as:
\begin{equation}
\mathcal{L}_{rec}=\| {T}-{\hat T}\|_1, \, \,
\mathcal{L}_{cyc}=\| {Z}_{tex}-{\hat Z}_{tex}\|_2,
\end{equation}
and the total loss for the texture network is formulated as follows:
\begin{equation}
\mathcal{L}_{T}=\mathcal{L}_{rec}+\mathcal{L}_{cyc} .
\label{eq:loss_texture_net}
\end{equation}

\subsection{Unified Expression Representation}

For each training frame $t$, our previously discussed three networks predict the expression latent code ${Z}_{e}$, the geometry latent code ${Z}_{g}$ and the texture latent code ${Z}_{tex}$ respectively. Intuitively, we can learn a mapping from expressions to geometries and textures, as we have disentangled head pose and viewpoint into separate latent codes.
We train an MLP $\mathbf{M}_e$ to learn mappings from the facial expression latent code to the geometry and texture latent codes. Given an expression ${Z}_{e}$ as input, $\mathbf{M}_e$ directly predicts ${\tilde Z}_{g}$ and ${\tilde Z}_{tex}$ as:
\begin{equation}
    {\tilde Z}_{g}, \, {\tilde Z}_{tex}=\mathbf{M}_e({Z}_{e}) .
\end{equation}
The mapping loss is straightforward which minimizes the distance between the encoded and mapped latent codes:
\begin{equation}
    \mathcal{L}_{map}=
    \|{Z}_{g}-{\tilde Z}_{g}\|_2+
    \|{Z}_{tex}-{\tilde Z}_{tex}\|_2.
\label{eq:mapping}
\end{equation}

\subsection{Training and Inference}

\vspace{3mm}\noindent{\bf Training.}  \label{sec:training}
Here we introduce an effective training scheme for our neural representation.
We first separately train our expression, geometry and texture VAE branches (Sec.~\ref{sec:4.4},~\ref{sec:4.5},~\ref{sec:4.6} respectively), to learn latent representations independently. 
After the three networks converge, we fix VAEs in all three branches and only train the latent mapping MLP $\mathbf{M}_e$.
Then, we apply an additional fine-tuning process to optimize the total loss. In this fine-tuning stage, we only lock the parameters of the expression network, and jointly train all other parts of the whole network (i.e., the geometry, texture, and mapping network). This strategy keeps the expression latent code stable and allows for more precise geometry and texture adjustment by fine-tuning the geometry and texture latent representations. Note that in the actual implementation, we minimize the Kullback-Leibler (KL) divergence loss for the latent codes in VAEs as well.

\vspace{3mm}\noindent{\bf Inference.}  \label{sec:inference}
Our neural facial asset representation unifies the person's specific expression, geometry, and texture latent codes. It grants us the ability to decode the geometry and physically-based texture assets directly from expressions.
At the inference stage, the input is a face image with certain expressions, we then generate the expression latent code using $\mathcal{E}_e$ and use $\mathbf{M}_e$ to translate the encoded expression latent code to geometry and texture latent codes. Subsequently, we use the geometry and texture decoders $\mathcal{D}_g$ and $\mathcal{D}_{tex}$ to generate the facial geometry, albedo, specular map, and normal map.
We formulate the inference process as: given an input image $\mathbf{I}$, we predict both the geometry $\mathbf{\tilde G}$ and the textures $\mathbf{\tilde T}$ with our expression encoder, latent mapping multilayer perceptron, and geometry and texture decoder:
\begin{equation}
\begin{aligned}
{Z}_{e}&=\mathcal{E}_{e} (I),\\
{\hat Z}_{g}, \, {\hat Z}_{tex} &=\mathbf{M}_e({Z}_{e}),\\
{\tilde G}&=\mathcal{D}_{g}({\hat Z}_{g}),\\
{\tilde W}&=\mathcal{D}_{tex}({\hat Z}_{tex}),\\
{\tilde T}&={\tilde W}\oplus \mathbf{N},
\end{aligned}
\end{equation}
where the viewpoint latent code and the geometry pose latent code are ignored since we intend to produce results that only depend on the facial expressions of the performer. Such pipeline produces consistent geometry and textures across different views of the input images due to our latent disentanglement network design.

\vspace{3mm}\noindent{}
{We provide the implementation details of our framework in our supplementary material. }
For further application of cross-identity neural retargeting, which will be mentioned in \ref{sec:5.2}, inspired by \citet{lombardi2018deep}, we propose a similar but different network design.
In our expression decoder, we concatenate a channel of zeros to each input before convolutional layers, denoted as \textbf{indicator channel}, which will be further used for cross-identity neural retargeting in \ref{sec:5.2}.

\begin{figure}[t]
    \centering
    \includegraphics[width=0.9\linewidth]{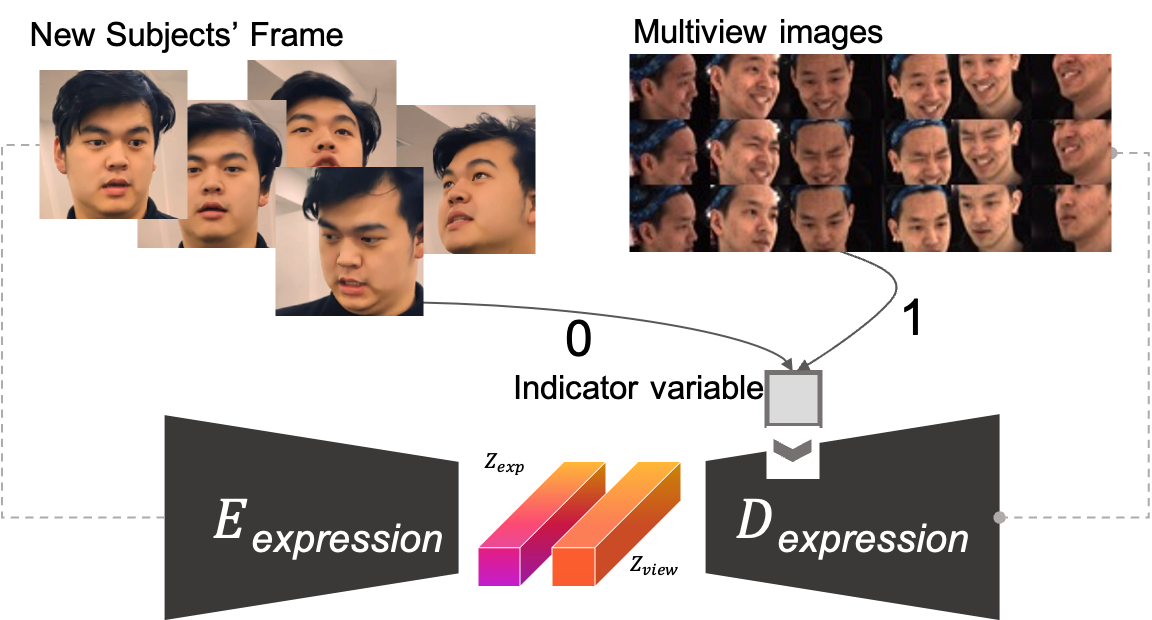}
    \caption{
\textbf{Indicator variable.}
The indicator variable informs the decoder of which domain it should decode the image into.
    }
    \label{fig:indicator}
\end{figure}

\section{Applications}
\label{sec:application}
Once our neural facial asset generator is trained for a specific performer, we can apply it in several scenarios with slight modification of our network. 
We can generalize our network to predict the performer's appearances with novel expressions, even animated by another person's in-the-wild face video. We also introduce various editing effects using our facial assets, including material editing, character animation and wrinkle transfer.

\subsection{Performer-specific neural facial asset}
Once trained,  our neural facial asset can be directly applied to produce unseen facial animations of the performer. It takes the monocular camera footage as input and outputs the geometry and corresponding multi-channel wrinkle maps, producing photo-realistic rendering results under novel facial animations from novel views and lighting. Thus, it can effectively enrich current scarce dynamic facial assets, and makes it possible to preview the dynamic facial assets instantly. Such strategy removes the process to capture and reconstruct unseen facial animations using the complex hardware setup.

\begin{figure}[t]
    \centering
    \includegraphics[width=0.9\linewidth]{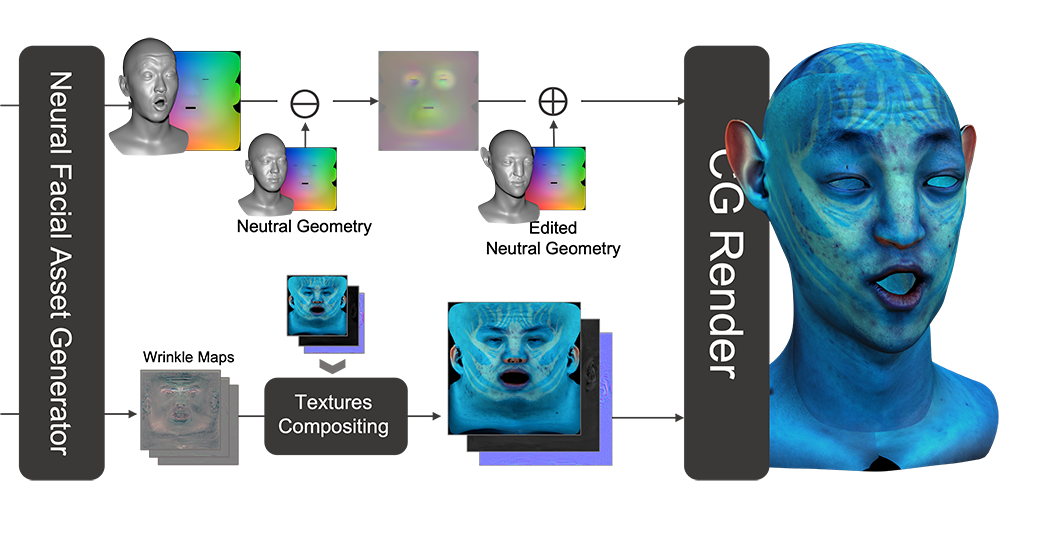}
    \caption{
\textbf{Illustration of geometry and texture editing pipeline.}
We can produce the realistic appearance of a novel character with consistent expressions.
    }
    \label{fig:texture_editing}
\end{figure}

\subsection{Cross-identity neural retargeting} \label{sec:5.2}
The movie and game industry has always had a keen interest in facial performance capture.
A common request from the industry is to drive the face animation of a digital character from the facial expressions of a real performer. Recall that our expression encoder can extract the performer's expressions, hence it only needs minor refinement when we apply it to another individual,
as presented in Fig.~\ref{fig:indicator}.

Specifically, given a well-trained neural physically-based facial asset, first, we extend the expression training dataset with footage of the new subject with different expressions and varying head poses. 
Then, we refine the expression network by training on this extended dataset.
Besides the losses mentioned in Eq.~\ref{eq:loss_expression_net}, we simultaneously minimize the reconstruction loss of images from the new subject with a strategy similar to the ones used in \cite{higgins2016beta} and \cite{lombardi2018deep}.
To let the expression decoder reconstruct images for the driving subject, we set the \textbf{indicator channel} of the decoder to 1 for images from footage of the driving subject and 0 for images from the original dataset.
This indicator allows the expression latent code to contain no identity-specific information but only expressions.
Moreover, to ensure the encoder produces the expression latent code independent from identity, we further add a discriminator {with RevGrad layer~\citet{pmlr-v37-ganin15}} to supervise the expression latent code.

Once trained the modified expression network, we lock its parameters and then refine both the geometry network and texture network together with the latent mapping MLP $\mathbf{M}_e$ on the original training dataset. 
Then, the new subject's performance can be transferred to the original performer.
The disentanglement of expression and viewpoint also allows the encoder to stably work with images taken under varying head poses, which means the new subject can even drive the performer with portable capture devices, such as cellphones. %

\begin{figure*}[t]
    \setlength{\abovecaptionskip}{5pt}
    \centering
    \setlength\tabcolsep{2pt}
    \renewcommand{\arraystretch}{0.1}
    
    \begin{tabular}{ccc}
         \includegraphics[width=0.32\linewidth]{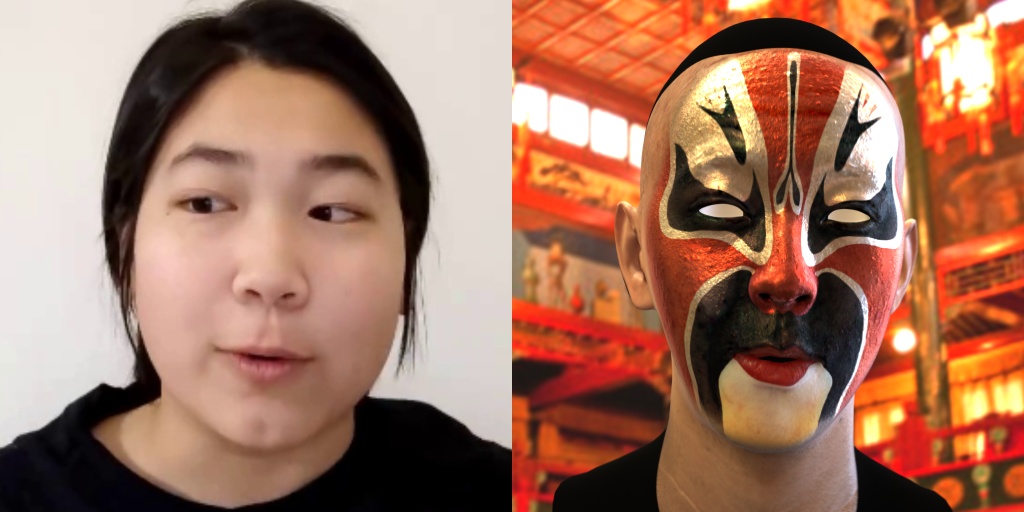}&\includegraphics[width=0.32\linewidth]{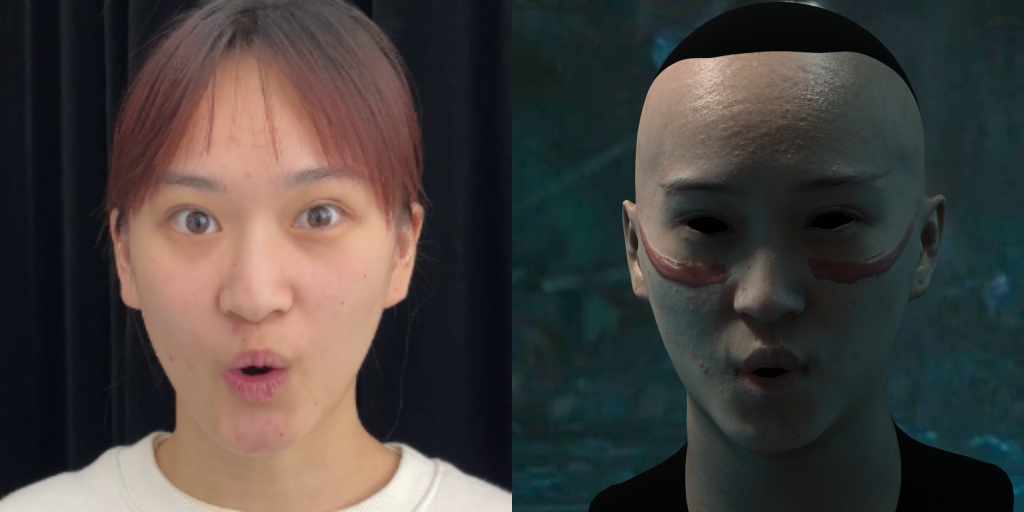} &\includegraphics[width=0.32\linewidth]{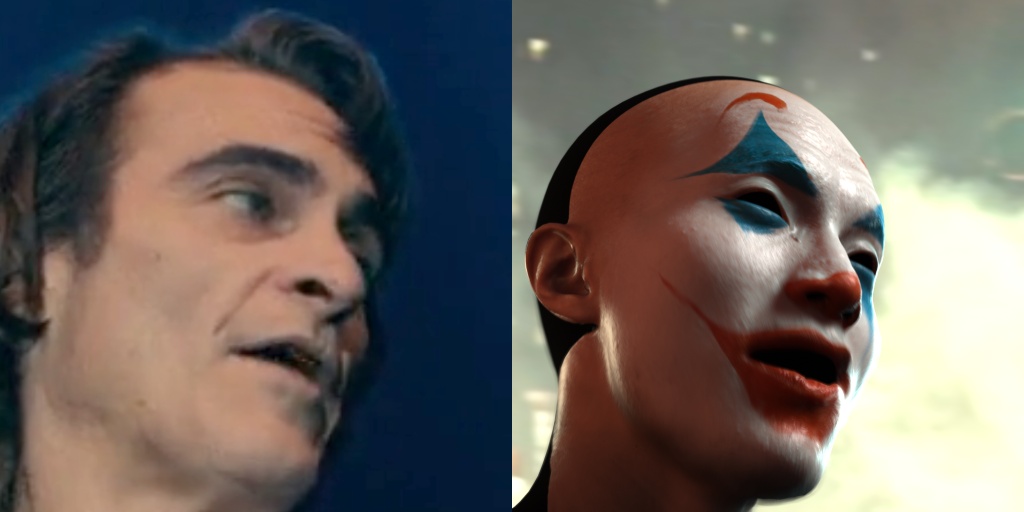}  \\
         \includegraphics[width=0.32\linewidth]{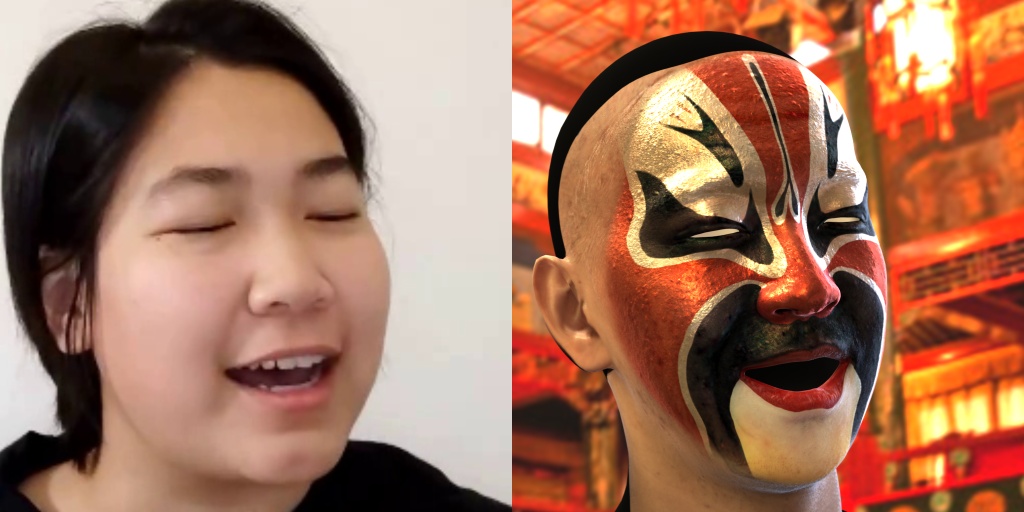}&\includegraphics[width=0.32\linewidth]{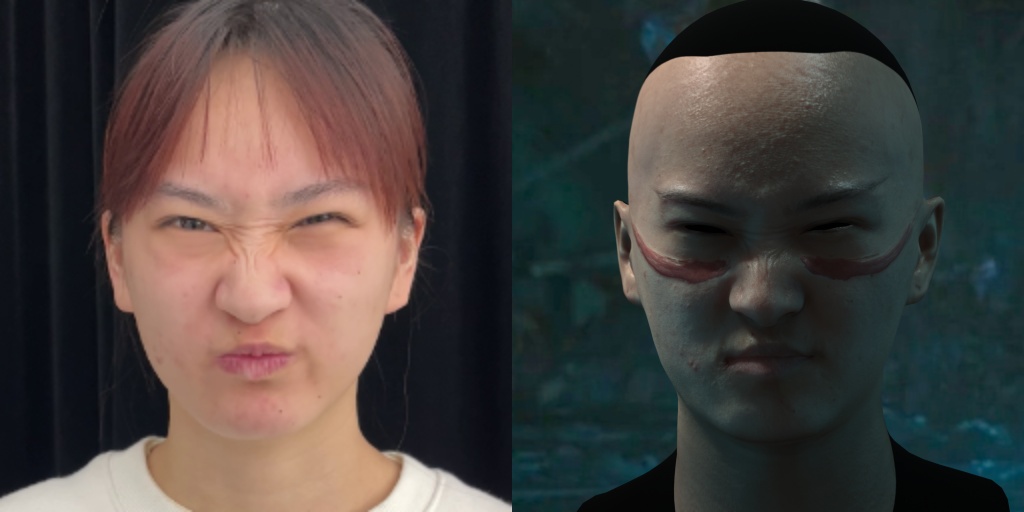} &\includegraphics[width=0.32\linewidth]{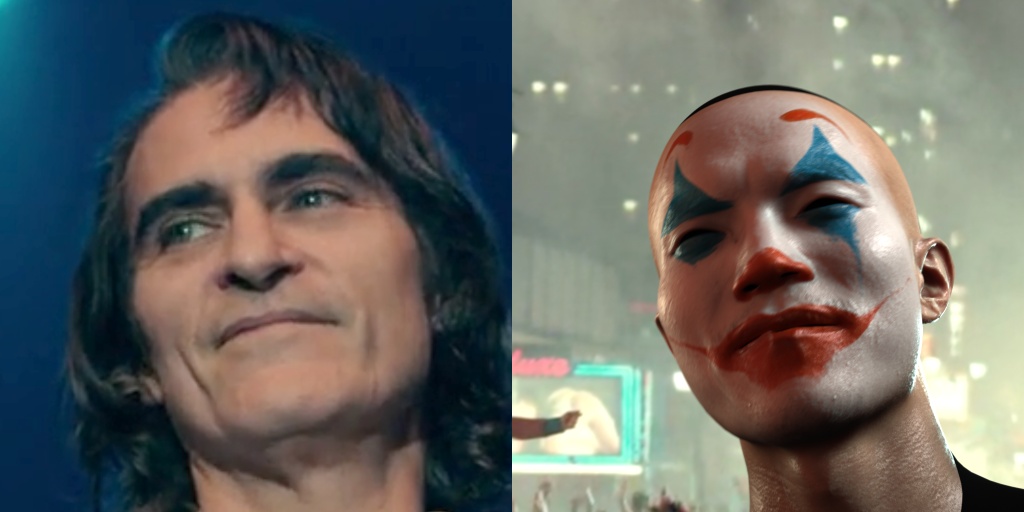}  \\
        \\
    \end{tabular}

    \caption{
\textbf{Re-targeting facial motions from videos using our neural physically-based facial assets with edited geometry and texture.}
{Each tuple includes driving frames and facial assets, with style of Guan Yu (left), Alita (middle), and Joker (right).}
    }
    \label{fig:gallery}
\end{figure*}

\subsection{Geometry and texture editing}
{Animation of computer augmented faces is essential for the movie and game industry, which involves geometry and texture editing of the captured face model and requires artist labor.
For instance, animating digital characters in the Avatar movie involves manual creation of novel appearances and hundreds of FACS action units and corresponding textures in order to generate realistic appearances, expressions and wrinkles.}
Previous works focus on high-precision facial capture or neural rendered final effects, yet they can not provide the artists with sufficient editability in textures, geometry, and shader to release the manual editing burden. 
In contrast, with our method, driving a stylized character becomes much easier, where only the neutral geometry and textures for the fictional character are needed.
The editing effort can be significantly reduced, as presented in Figure \ref{fig:texture_editing}.
To drive the stylized character, we first create the character under neutral expression with artificial facial details and novel skin appearances. 
We can modify the geometry to change facial features, e.g. longer ears and thinner cheeks.
Similarly, we can modify the textures by painting on diffuse albedo, changing micro-geometry on normal map and editing specular intensity.
We expect this new character to exhibit similar movements in facial muscle and wrinkle details.
Then, we transfer the dynamic facial geometry as well as wrinkle maps under different expressions to drive this new character.

{Taking the driving subject's facial performance video as input, our neural facial asset can predict per-frame dynamic geometry. 
We calculate the facial movement, which we regard as an offset field, and then apply the offset to the neutral geometry map of stylized character to obtain the geometry with the same expression.
During the texture compositing stage, we replace the original neutral textures with an exquisitely drawn one.
Because our network is designed to predict wrinkle maps, it can transfer the performer's wrinkles to the stylized character. 
}

\section{Results}

In this section, we demonstrate the capability of our approach in a variety of scenarios.
We first report the implementation details of our approach and the dataset captured by our FaStage, and then analyze our results in various editing scenarios. 
We further provide a comparison with previous methods and evaluations of our main technical components, both qualitatively and quantitatively. 
The limitation and discussions regarding our approach are provided in the last subsection.

\paragraph{Dataset and Implementation Details.} 
To train and evaluate our method, we capture a dynamic facial asset dataset with physically-based textures using our FaStage system.
Our capture system is shown in Fig.~\ref{fig:data_acquisition}, where all the cameras are calibrated and synchronized with lighting in advance.
During capture, the machine vision camera array produces 24 RGB streams at 2592$\times$2048 resolution, while the high-speed camera array generates 3 RGB streams at 4096$\times$2304 resolution.
After data pre-processing, we produce 4096x4096 unwrapped physically-based textures for each frame, including diffuse albedo, normal map, and specular intensity.
Our dataset consists of three performers and we capture a total number of 7200 frames for each performer, respectively. 
{In our implementation of facial motion re-targeting, we use the first 30 seconds of the in-the-wild video clip for refinement.
We crop the frames in the clip to 256x256 with faces in the center using the landmark detector from Dlib~\cite{king2009dlib}, and only compute reconstruction loss within the face region.}
Additionally, we adopt Blender's Eevee as our main render engine as it is free, open-source, fast, and convenient. Rendering a frame with textures of 4K resolution takes less than 1 second with the Eevee engine. We further utilize the rendering shader provided in the Heretic ~\cite{Heretic} to achieve real-time performance.
Fig.~\ref{fig:gallery} demonstrates several neural facial assets generated using our approach for high-realistic facial capture and physically-based editing.

\begin{figure*}[t]
    \setlength{\abovecaptionskip}{0pt}
    \centering
    \includegraphics[width=\linewidth]{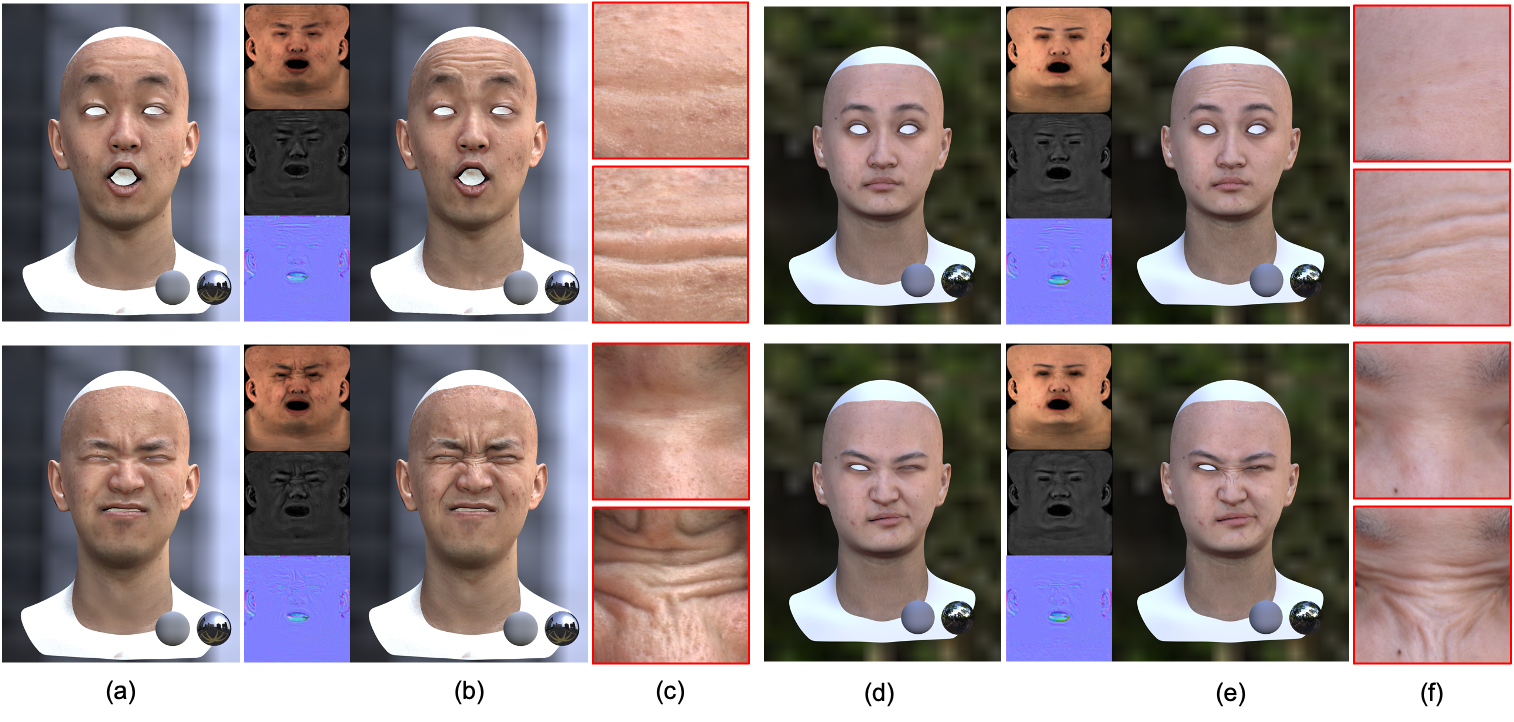}
    \caption{
\textbf{Video-driven Neural Physically-based Facial Asset.}
Our dynamic textures effectively enhance the appearance of the performer:
(a,d) driven facial assets with only static neutral physically-based textures,
(b,e) driven facial assets with dynamic physically-based textures,
(c,f) zoom-in view.
Our method successfully models the dynamic textures and preserves facial details at high resolution.
}
    \label{fig:application_gallery}
\end{figure*}

\begin{figure*}[h]
    \centering
    \includegraphics[width=\linewidth]{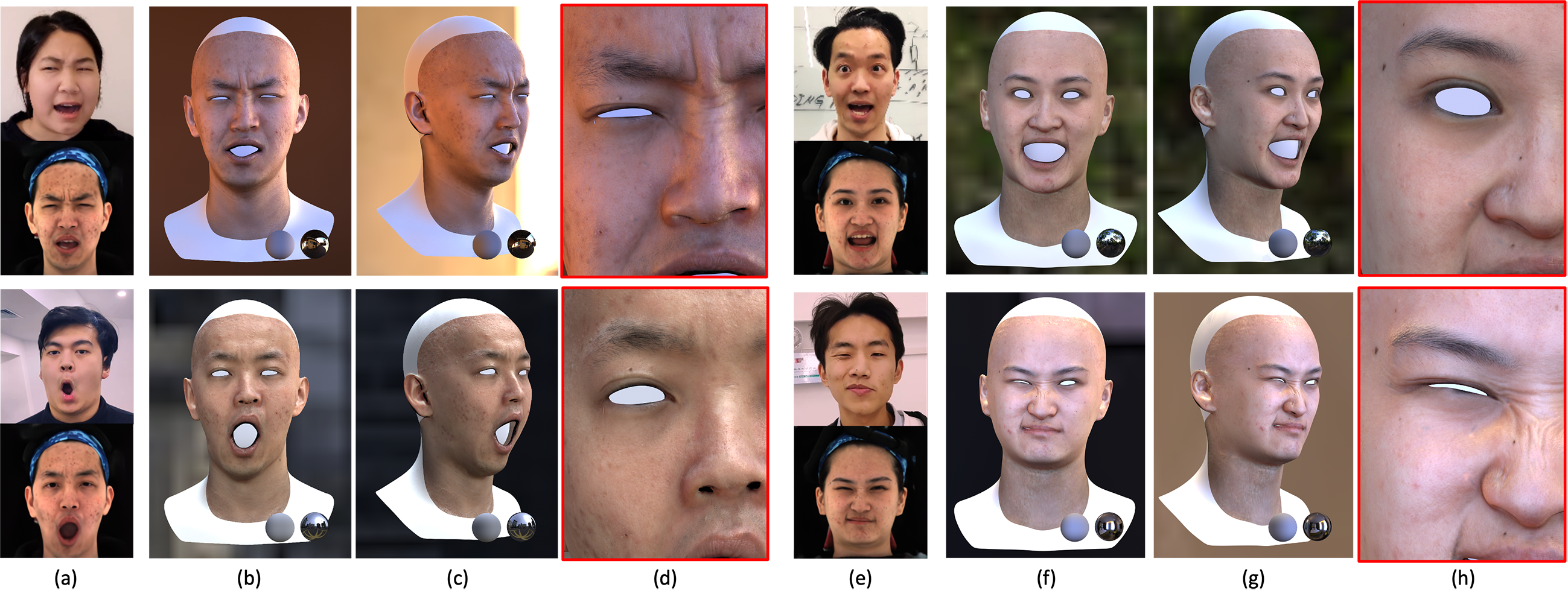}
    \caption{
\textbf{Cross-identity video driven results.}
We extend the expression network for new subjects with different expressions and varying head poses: 
(a,e) the upper image is the new performer's expression, and the lower image is the neural retargeting results from our expression decoder, 
(b,f) driven facial asset in front view, 
(c,g) driven facial asset in the left view, 
(d,h) zoom-in view. 
Our method achieves detailed video-driven results from different identities with dynamic textures, which leads to photo-realistic rendering.
}
    \label{fig:application_gallery1}
\end{figure*}

\begin{figure*}[h]
    \centering
    \includegraphics[width=\linewidth]{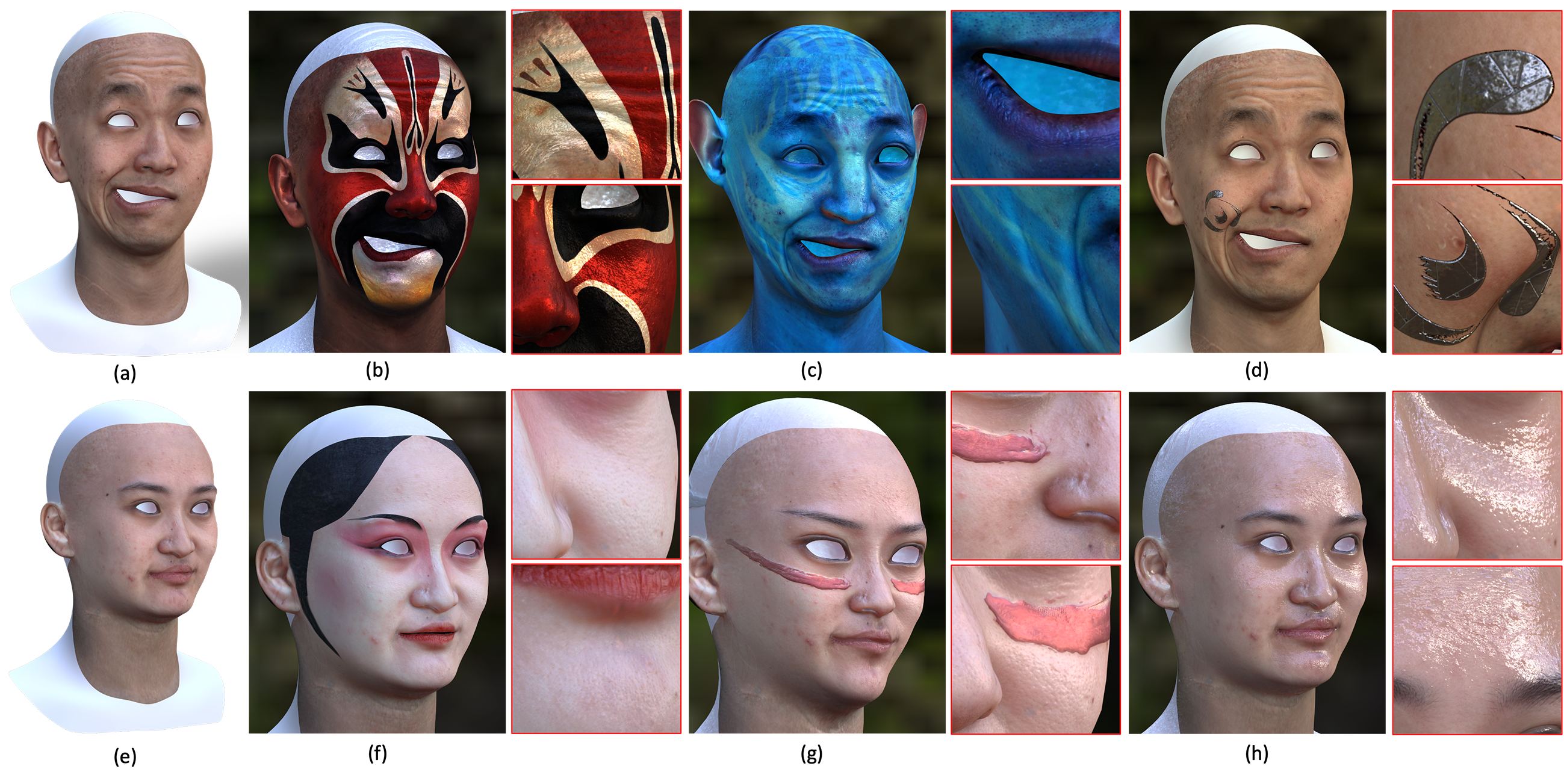}
    \caption{
\textbf{Geometry and texture editing results.}
We edit the video-driven facial assets (a,e) in various ways:
(b) We paint the diffuse albedo with the style of Guan Yu, the most famous red-face character in traditional Chinese Peking Opera;
(c) We stylize the geometry of the performer to the blue alien from the feature film AVATAR with a novel facial structure but consistent identity features;
(d) We add a metal DEEMOS logo to the cheek of the performer by modifying the textures;
(f) We paint the diffuse albedo with the style of Yu Ji, the beloved concubine of Xiang Yu, the hegemon of Western Chu, from the famous Peking Opera \textit{Farewell My Concubine};
(g) We alter the performer's facial structure, and added realistic face painting by jointly modifying the diffuse albedo and the normal map;
(g) We adjust the facial roughness to give the skin a more shiny look.
}
    \label{fig:application_gallery2}
\end{figure*}

\begin{figure*}[t]
    \setlength{\abovecaptionskip}{-5pt}
    \centering
    \includegraphics[width=\linewidth]{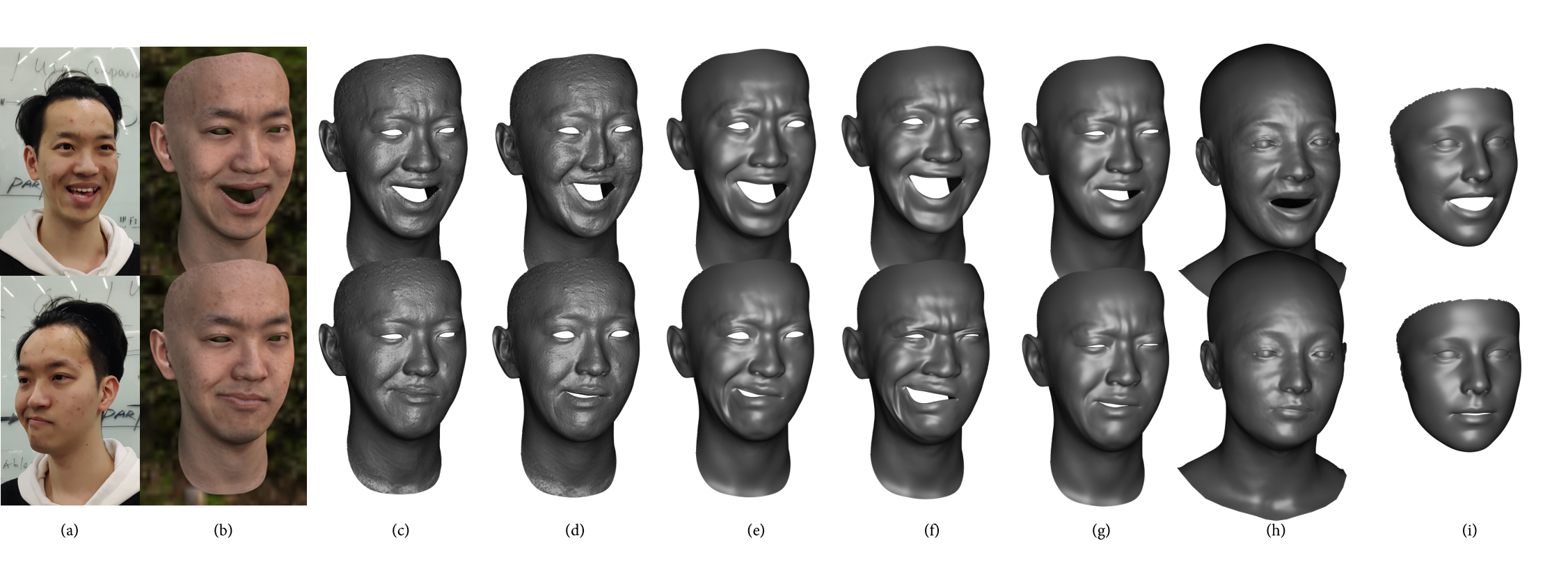}
    \caption{
\textbf{Qualitative comparison of video-driven result on in-the-wild monocular RGB inputs.}
(a) the input in-the-wild image,
(b) rendered result of our method with textures,
(c) predicted geometry of our method with normal map, after the refinement using in-the-wild images,
(d) predicted geometry of our method with normal map, before the refinement using in-the-wild images,
(e) predicted geometry of \citet{lombardi2018deep}, trained with in-the-wild images,
(f) predicted geometry of \citet{lombardi2018deep}, trained without in-the-wild images,
(g) predicted geometry of \citet{laine2017production},
(h) predicted geometry of \citet{DECA:Siggraph2021},
(i) predicted geometry of \citet{guo2020towards}.
With a quick refinement, our method can deal with head pose and lighting variance better than other methods.
    }
    \label{fig:comparison_qual2}
\end{figure*}

\subsection{Video-driven Applications on Novel Expressions}
Our multi-VAE-based neural facial assets enable various video-driven applications, ranging from fine-grained facial capture to cross-identity facial motion retargeting, and to geometry and physically-based texture editing.  
As shown in Fig.~\ref{fig:application_gallery}, our approach enables high-quality dynamic facial geometry generation with pore-level dynamic textures on complicated expressions of the specific performer, such as frowning or winking. 
Fig.~\ref{fig:application_gallery1} further provides the video-driven results using the RGB footage of different performers. 
In addition to driving the captured performer, our method further enables convenient physically-based editing for generating and driving an artistically stylized avatar, {as shown in Fig.~\ref{fig:application_gallery2}.}
Besides, our retargeted animations are temporally consistent, which is provided in the accompanying video. 

\subsection{Comparisons}
To demonstrate the video-driven performance of our approach, we compare against various video-driven facial reconstruction and animation methods, including parametric facial model 3DDFA proposed by \citet{guo2020towards} and DECA proposed by \citet{DECA:Siggraph2021}, a performer-specific method for production proposed by \citet{laine2017production}, %
and Deep Appearance Model proposed by \citet{lombardi2018deep}.
We adopt the official pre-trained PyTorch model for the method of \citet{DECA:Siggraph2021} and \citet{guo2020towards}, and faithfully re-implement both the methods of \citet{laine2017production} and \citet{lombardi2018deep} and train their models using the same training dataset as ours for a fair comparison.

\begin{figure}[t]
    \centering
    \includegraphics[width=\linewidth]{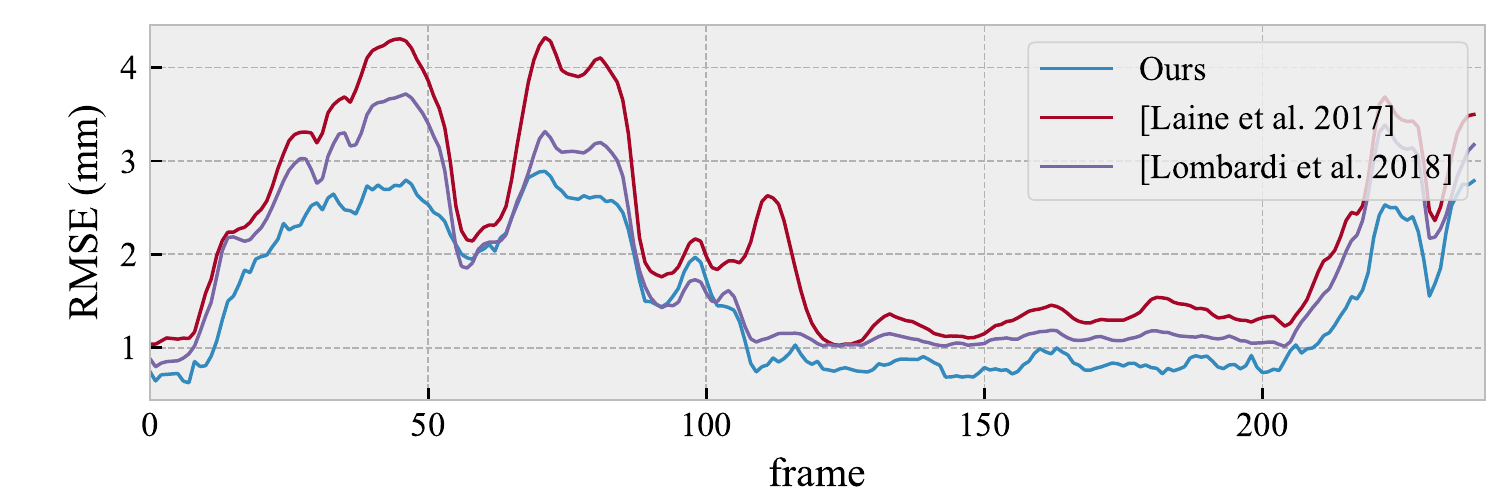}
    
    \vspace{10pt}
    
    \begin{tabular}{cccc}
    \hline\hline
        \multirow{2}{*}{Method} &\multicolumn{3}{c}{RMSE (mm)} \\
        &total&eye&mouth\\
        \hline\hline
        \citet{laine2017production} & 2.237&1.565& 4.157\\\hline
        \citet{lombardi2018deep} & 1.866 &  1.174&3.076\\\hline
        \textbf{Ours} & \textbf{1.516} & \textbf{1.164} & \textbf{2.750} \\\hline
    \end{tabular}
    \caption{
\textbf{Quantitative comparison against several state-of-the-art methods.}
We compare the per-frame geometric reconstruction errors on an image sequence in our testing dataset (upper) and overall geometric reconstruction errors on the images from our testing dataset (lower).
Our method outperforms others, indicating the better expression ability of our network.
    }
    \label{fig:comparison_time}
\end{figure}

We divide $20\%$ of the total frames of an individual as our testing set. We provide multi-view images as input for both the methods of \citet{laine2017production} and \citet{lombardi2018deep} in the training process.
Fig.~\ref{fig:comparison_qual2} provides the qualitative comparison of our testing dataset for the task of facial geometry capture from in-the-wild monocular video input.
It is worth noting that compared to other methods, our approach generates more realistic pore-level facial details which are exhibited in normal maps.
Besides, our approach generates more realistic video-driven facial animation results even under challenging expressions, which compares favorably to other methods.
For fair quantitative comparison, we first transform the output models on our testing dataset from both \citet{laine2017production} and \citet{lombardi2018deep}  into our geometry map representation.
Note that the unit of coordinates is transformed into millimeters and we adopt RMSE as the metric. 
As shown in Fig.~\ref{fig:comparison_time}, our approach consistently outperforms the baselines, which demonstrates the effectiveness of our approach for dynamic video-driven facial capture and modeling.
{More importantly, our approach enables various physically-based applications like geometry and material editing or facial feature transfer, which are rare in the above methods.}

\begin{figure}
    \centering
    \setlength\tabcolsep{1pt}
    \begin{tabular}{ccc}
          \rotatebox[origin=c]{90}{\small{frame 525, 526}}& \raisebox{-0.5\height}{\includegraphics[width=0.4\linewidth]{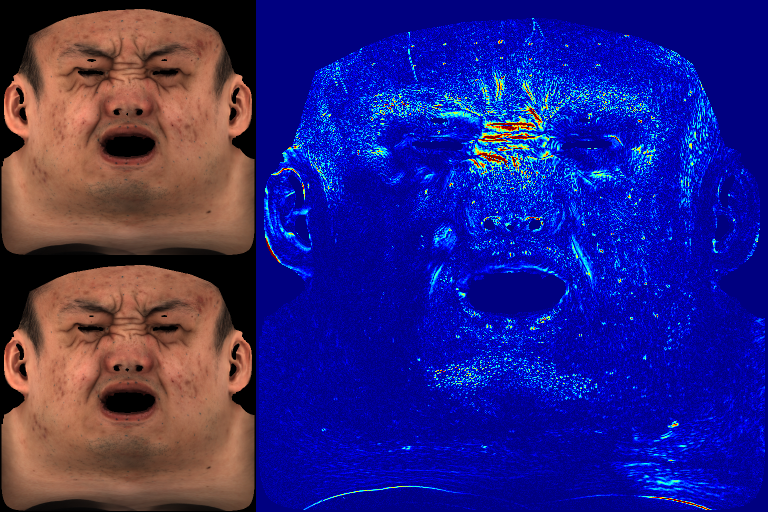}} & \raisebox{-0.5\height}{\includegraphics[width=0.4\linewidth]{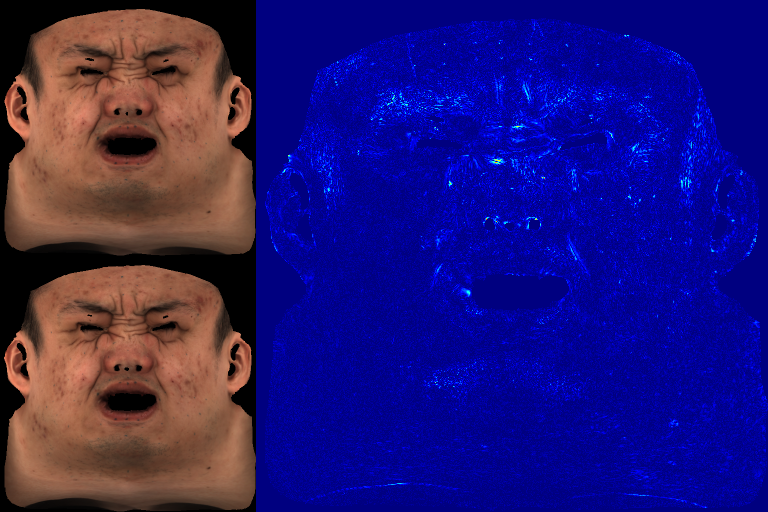}}\vspace{4pt} \\
          \rotatebox[origin=c]{90}{\small{frame 1036, 1037}}&
          \raisebox{-0.5\height}{\includegraphics[width=0.4\linewidth]{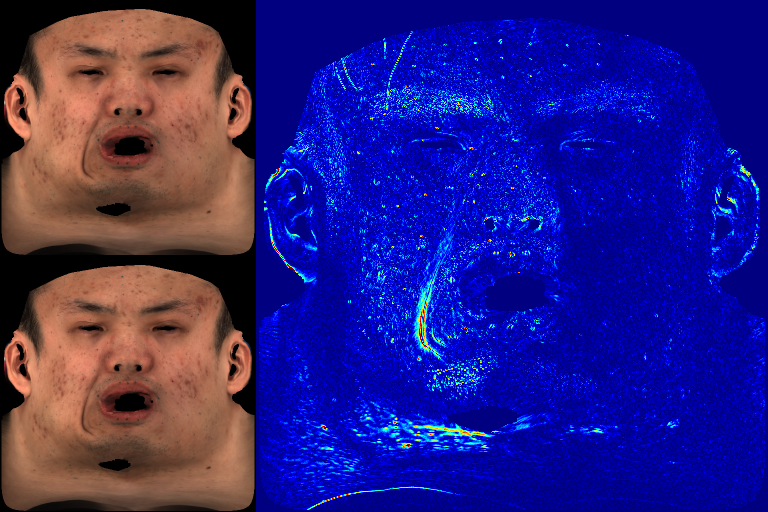}} & \raisebox{-0.5\height}{\includegraphics[width=0.4\linewidth]{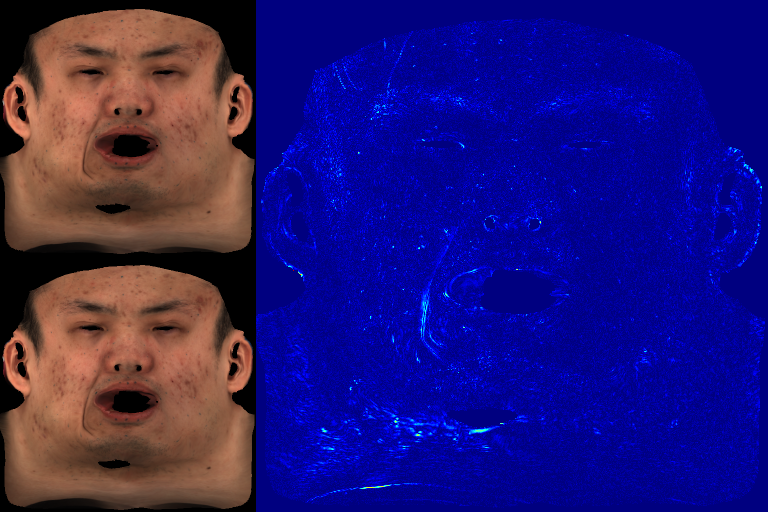}} \\
          &w/o stabilization & Ours
    \end{tabular}
    \includegraphics[width=1\linewidth]{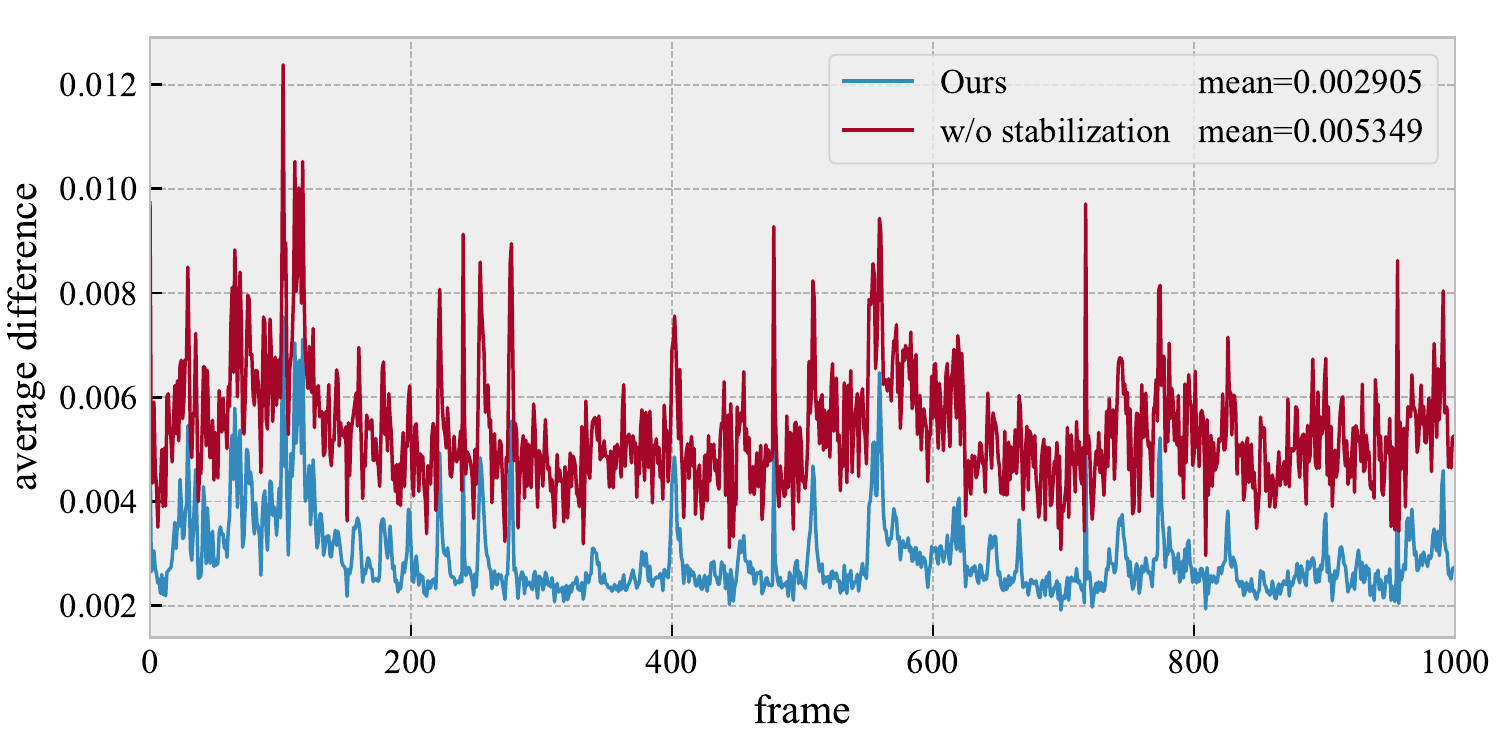}
    \caption{\textbf{Effectiveness of temporal stabilization.} The upper figures show the per-pixel difference of diffuse textures from consecutive frames with and without temporal stabilization. The lower curve shows the average difference of consecutive diffuse textures in our first 1000 frames. With temporal stabilization applied, we significantly reduce the difference between consecutive frames and provide more consistent textures.}
    \label{fig:comparison_temporal}
\end{figure}

\subsection{Ablation Study}
Here, we evaluate the performance of different training strategies and demonstrate the effectiveness of our multi-VAE network designs.
Specifically, we evaluate the temporal stabilization, the expression disentanglement, the pose enhancement, and wrinkle-based texture prediction in our expression, geometry, and texture VAE, respectively.

\paragraph{Temporal stabilization.}
We compare the stability of our textures with or without the temporal stabilization.
Let \textbf{w/o stabilization} denote the results of our original textures without temporal stabilization.
We compare the difference of textures between consecutive frames, where the lower difference indicates less temporal instability.
As illustrated in Fig.~\ref{fig:comparison_temporal}, our temporal stabilization significantly improves the stability of textures for later network training.

\begin{figure}[t]
    \centering
    \includegraphics[width=\linewidth]{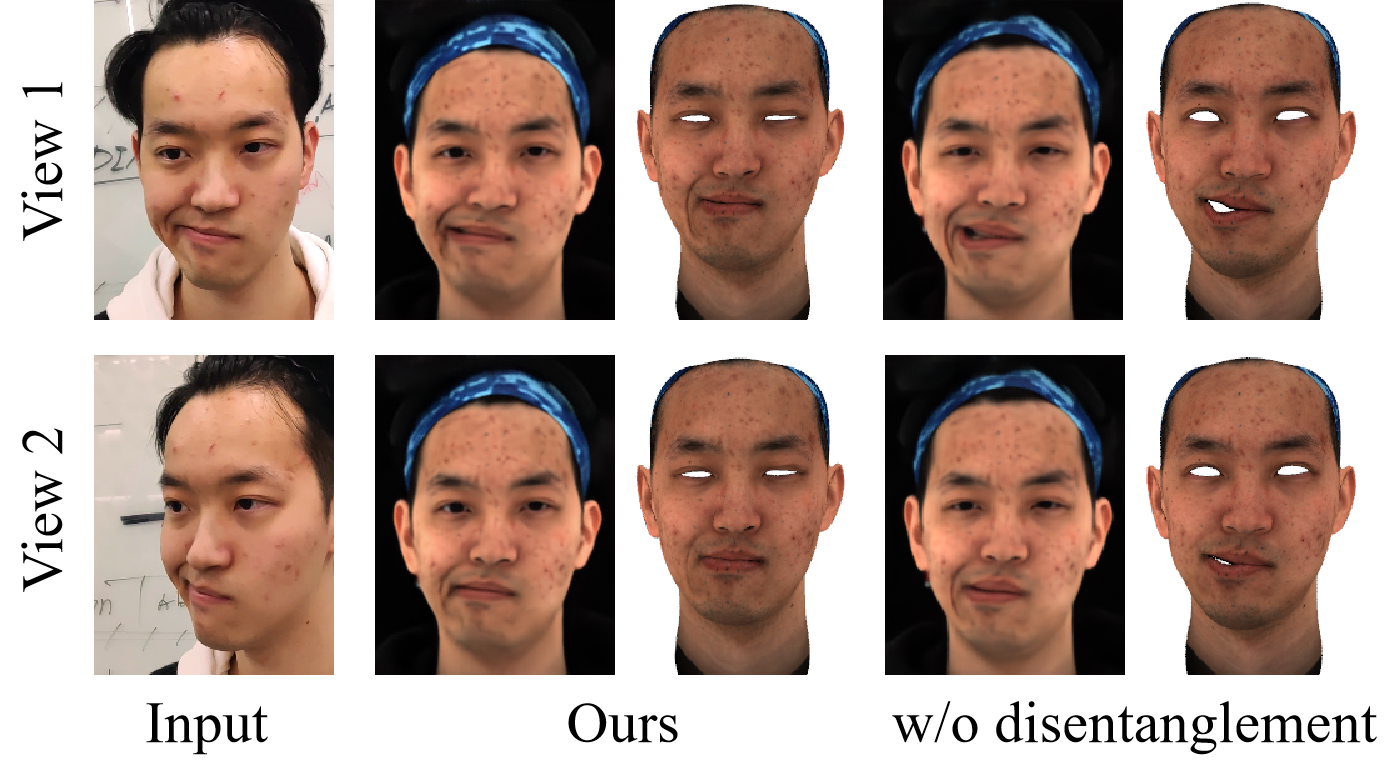}
    \caption{
\textbf{Qualitative evaluation of our expression disentanglement.}
After training the network with the same dataset, we tested it with images from novel viewpoints.
The variation without expression disentanglement suffers from inconsistency when the viewpoint changes.
The absence of expression and viewpoint disentanglement leads to a reduced ability of generalization for our model. 
As a comparison, the results given by our model with expression disentanglement training handled viewpoint change with no deterioration in performance. 
    }
    \label{fig:ablation_disentanglement}
\end{figure}

\paragraph{Expression disentanglement.}
We compare the performance of our model with or without the expression disentanglement training scheme.
Let \textbf{w/o disentanglement} denote the variation of our approach without triplet input for both expression and geometry VAEs. 
We utilize the same multi-view images as input for training but ignore the viewpoint information.
During the training process of the expression network, we only adopt the loss $\mathcal{L}_{rec}$ in Eq.~\ref{eq:Lrecon}, and remove viewpoint latent code from our network.
As shown in Fig.~\ref{fig:ablation_disentanglement}, our scheme with expression disentanglement enables more accurate and consistent prediction for geometry, especially when the input viewpoint changes.
Our expression disentanglement training scheme facilitates the robustness of our model to handle viewpoint variance.

\begin{figure}[t]
    \centering
    \includegraphics[width=1\linewidth]{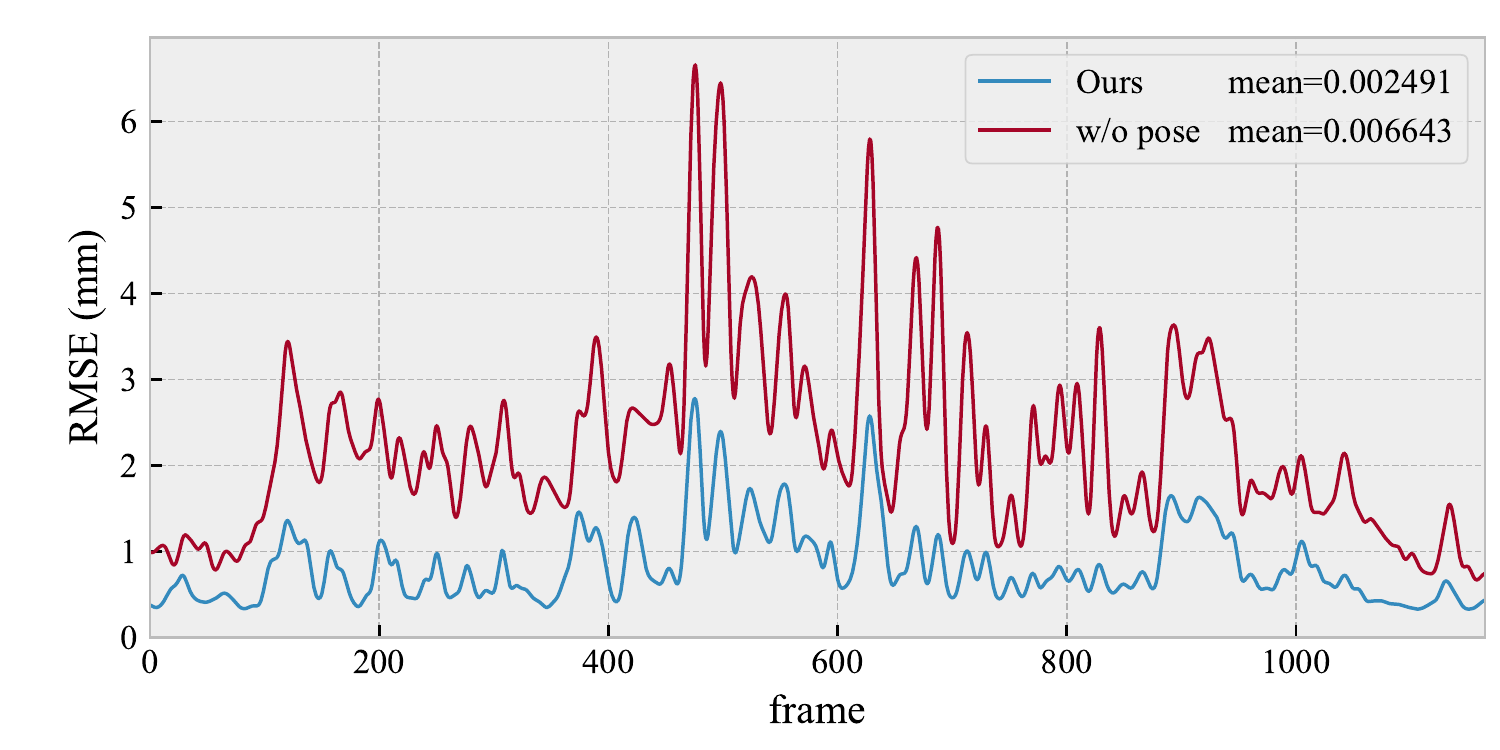}
    \caption{
\textbf{Quantitative evaluation of our pose augmentation.}
We evaluate the stability of the predicted geometries with and without pose augmentations.
The curve shows the mean distances of vertices to their stable coordinates, which are expected to be fixed.
Our training scheme with pose augmentation leads to more stable face geometries, while the results given by the variation without it jitters while testing.
    }
    \label{fig:ablation_pose}
    \vspace{-0.3cm}
\end{figure}

\paragraph{Pose extraction.}
Here we compare against our variation without encoding the pose latent code from geometry while keeping the input geometry intact without augmentation, denoted as \textbf{w/o pose}.
During the training process, we only adopt the first two loss items in Eq.~\ref{eq:loss_geometry_net}.
As illustrated in Fig.~\ref{fig:ablation_pose}, the results of \textbf{w/o pose} are temporally unstable. 
We observe several sudden changes in geometry, leading to the spikes of the curves in Fig.~\ref{fig:ablation_pose}.
Such an unstable phenomenon compromises the performance of facial animation, causing manual correction in post-production.
Differently, our pose enhancement handles the head pose variance and alleviates the unstable artifact.

\begin{figure}[t]
    \centering
    \includegraphics[width=0.9\linewidth]{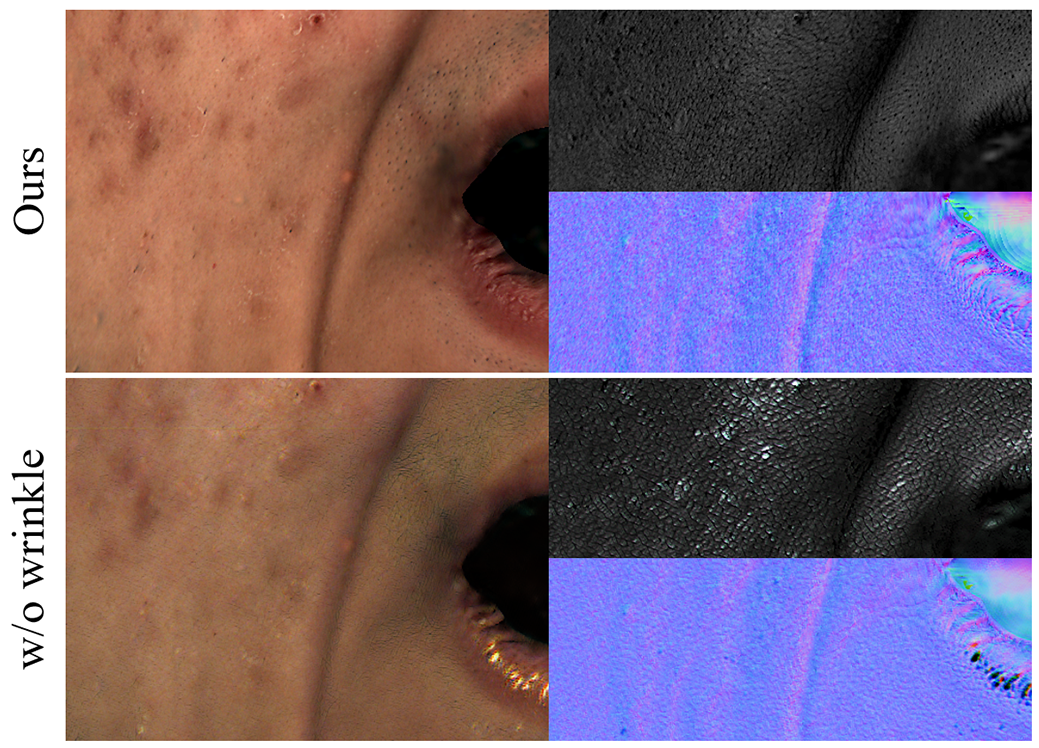}
    \caption{
\textbf{Qualitative evaluation of our texture predicting strategy.}
Our variation without predicting wrinkles suffers from repeated artifacts and loses a lot of real details. 
The wrinkle map preserves more facial details and sharpens the textures {and avoids} scale-like artifact.
    }
    \label{fig:ablation_texture}
\end{figure}

\paragraph{Texture prediction strategy.}
We further evaluate our wrinkle-map-based texture prediction strategy.
Let \textbf{w/o wrinkle} denote our variation that directly predicts pixel-aligned textures instead of wrinkle maps and subsequently adopts an extra super-resolution network to upscale the predicted textures to 4K resolution.
As shown in Fig.~\ref{fig:ablation_texture}, our scheme models the wrinkle maps including shading maps, which enables better preservation of facial details.
Our wrinkle map models the discrepancies on top of the neutral textures and can be scaled to high-resolution without suffering from blur or block-wise artifacts.

\subsection{Limitations and Discussions}
{Though our neural physically-based facial asset has demonstrated its ability in a variety of applications with high-fidelity rendering performance, it is still subject to several limitations.
Our approach lacks semantic animation controls. We choose implicit representation to model the expressions over explicit parametric methods, therefore the re-targeting results of the asset cannot be easily edited by an artist. While this remains a limitation to our method, there are alternatives: one can either replace our in-the-wild data with rendered images from parametric models analogous to \cite{moser2021semi} or directly map extracted explicit parameters from the in-the-wild images to our latent variables. 
Our approach achieves cross-identity facial motion re-targeting with robust expression disentanglement. Yet we capture the facial asset in a performer-specific paradigm, hence it can not well handle extremely challenging identity and environmental variance, similar to other performer-specific work~\cite{laine2017production,lombardi2018deep}.}
Besides, our cross-identity facial motion retargeting relies on actor-specific training when applying to a new performer. %
It's an interesting direction to further design a generalizable expression network structure to simultaneously handle various identities, similar to those general facial models~\cite{DECA:Siggraph2021}.
Moreover, we don't simulate the lighting conditions and background as well as previous methods (i.e. \citet{klaudiny2017real}).
In future works, we plan to apply single-image relighting techniques with portrait matting (i.e. \citet{zhang2021neural}) to simulate novel lighting environments. 

Our geometry network only models soft tissues of the face with high fidelity, we still need other face components like teeth, gums, eyeballs, eye occlusion, and eyelashes, to improve the realism of our facial assets. Previous works apply different types of deformations, such as linear and laplacian deformations, to model {the} movements of these components, which may not be able to achieve high realism. In future works, we will incorporate eye movement capture and jaw adaptation algorithms into our model, and use a biologically sound parametric skull model as prior.
Apart from face components, we also need realistic hair with correct deformations to build a production-level digital avatar. Rendering hair with high fidelity remains a challenging problem in computer graphics due to its complex physical forms. Existing solutions are mostly computationally expensive, yet recent work using neural radiance field ~\cite{mildenhall2020nerf} to render fuzzy objects has shown promising results. Hence, we plan to further improve our method by rendering hair simultaneously using the method of~\cite{luo2021convolutional}.

In our practice, we separate textures into neutral textures and multi-channel wrinkle maps for modifying high-resolution textures. They can not simulate certain phenomena, such as pore squeezing. 
A possible solution is to apply a similar strategy as we use at the data processing stage, where we disentangle textures into warp fields and neutral textures to eliminate pixel drifts for alignment.
We have tried such approach for modeling skin stretch and compression when generating textures, but failed due to jittering noise in the data.
In future works, we will further improve our data processing pipeline, which upgrades the current wrinkle-map-based texture inference framework to a warp-field-based one. In the new framework, the dynamic texture generated by our network will be able to simulate pore-level deformation effects.

\section{Conclusion}
We have presented a neural approach to generate video-driven dynamic facial geometries with high-quality physically-based texture assets from FaStage data. Our neural approach enables high-fidelity performer-specific facial capture and even cross-identity facial motion re-targeting without manual labor. It also enables various physically-based facial editing effects, like geometry and material editing or wrinkle transfer.
The core of our approach is a new multi-VAE-based design to disentangle facial attributes explicitly and effectively with the aid of a global MLP-based mapping across latent spaces. Our explicit expression and viewpoint disentanglement as well as the novel triplet semi-supervision scheme enable effective latent space construction for both the expression and geometry VAEs. Our wrinkle maps and view-independent texture inference in our physically-based texture VAE achieve high-quality 4K rendering of dynamic textures and convenient neutral-based editing for post-production.  
We showcase extensive experimental results and video-driven applications for {high-fidelity} facial animation, facial motion retargeting, and physically-based editing with high realism. We believe that our approach significantly improves the video-driven production-level workflow of generating dynamic facial geometries with physically-based assets. It serves as a critical step for automatic, high-quality, and controllable dynamic facial asset generation in the neural era, with many potential applications for believable digital human, film, game production, and immersive experience in VR/AR and the Metaverse.

\begin{acks}

This work was supported by Shanghai YangFan Program (21YF1429500), Shanghai Local College Capacity Building Program (22010502800), NSFC programs (61976138, 61977047), the National Key Research and Development Program (2018YFB2100500), STCSM (2015F0203-000-06) and SHMEC (2019-01-07-00-01-E00003).

\end{acks}

\bibliographystyle{ACM-Reference-Format}
\bibliography{sample-bibliography}


\begin{thebibliography}{84}


\ifx \showCODEN    \undefined \def \showCODEN     #1{\unskip}     \fi
\ifx \showDOI      \undefined \def \showDOI       #1{#1}\fi
\ifx \showISBNx    \undefined \def \showISBNx     #1{\unskip}     \fi
\ifx \showISBNxiii \undefined \def \showISBNxiii  #1{\unskip}     \fi
\ifx \showISSN     \undefined \def \showISSN      #1{\unskip}     \fi
\ifx \showLCCN     \undefined \def \showLCCN      #1{\unskip}     \fi
\ifx \shownote     \undefined \def \shownote      #1{#1}          \fi
\ifx \showarticletitle \undefined \def \showarticletitle #1{#1}   \fi
\ifx \showURL      \undefined \def \showURL       {\relax}        \fi
\providecommand\bibfield[2]{#2}
\providecommand\bibinfo[2]{#2}
\providecommand\natexlab[1]{#1}
\providecommand\showeprint[2][]{arXiv:#2}

\bibitem[Aberman et~al\mbox{.}(2019)]%
        {aberman2019learning}
\bibfield{author}{\bibinfo{person}{Kfir Aberman}, \bibinfo{person}{Rundi Wu},
  \bibinfo{person}{Dani Lischinski}, \bibinfo{person}{Baoquan Chen}, {and}
  \bibinfo{person}{Daniel Cohen-Or}.} \bibinfo{year}{2019}\natexlab{}.
\newblock \showarticletitle{Learning Character-Agnostic Motion for Motion
  Retargeting in 2D}.
\newblock \bibinfo{journal}{\emph{ACM Transactions on Graphics (TOG)}}
  \bibinfo{volume}{38}, \bibinfo{number}{4} (\bibinfo{year}{2019}),
  \bibinfo{pages}{75}.
\newblock


\bibitem[Abrevaya et~al\mbox{.}(2020)]%
        {abrevaya2020cross}
\bibfield{author}{\bibinfo{person}{Victoria~Fern{\'a}ndez Abrevaya},
  \bibinfo{person}{Adnane Boukhayma}, \bibinfo{person}{Philip~HS Torr}, {and}
  \bibinfo{person}{Edmond Boyer}.} \bibinfo{year}{2020}\natexlab{}.
\newblock \showarticletitle{Cross-modal deep face normals with deactivable skip
  connections}. In \bibinfo{booktitle}{\emph{Proceedings of the IEEE/CVF
  Conference on Computer Vision and Pattern Recognition}}.
  \bibinfo{pages}{4979--4989}.
\newblock


\bibitem[Bao et~al\mbox{.}(2021)]%
        {bao2020high}
\bibfield{author}{\bibinfo{person}{Linchao Bao}, \bibinfo{person}{Xiangkai
  Lin}, \bibinfo{person}{Yajing Chen}, \bibinfo{person}{Haoxian Zhang},
  \bibinfo{person}{Sheng Wang}, \bibinfo{person}{Xuefei Zhe},
  \bibinfo{person}{Di Kang}, \bibinfo{person}{Haozhi Huang},
  \bibinfo{person}{Xinwei Jiang}, \bibinfo{person}{Jue Wang}, {et~al\mbox{.}}}
  \bibinfo{year}{2021}\natexlab{}.
\newblock \showarticletitle{High-Fidelity 3D Digital Human Head Creation from
  RGB-D Selfies}.
\newblock \bibinfo{journal}{\emph{ACM Transactions on Graphics (TOG)}}
  \bibinfo{volume}{41}, \bibinfo{number}{1} (\bibinfo{year}{2021}),
  \bibinfo{pages}{1--21}.
\newblock


\bibitem[Beeler et~al\mbox{.}(2010)]%
        {beeler2010high}
\bibfield{author}{\bibinfo{person}{Thabo Beeler}, \bibinfo{person}{Bernd
  Bickel}, \bibinfo{person}{Paul Beardsley}, \bibinfo{person}{Bob Sumner},
  {and} \bibinfo{person}{Markus Gross}.} \bibinfo{year}{2010}\natexlab{}.
\newblock \showarticletitle{High-quality single-shot capture of facial
  geometry}.
\newblock In \bibinfo{booktitle}{\emph{ACM SIGGRAPH 2010 papers}}.
  \bibinfo{pages}{1--9}.
\newblock


\bibitem[Bi et~al\mbox{.}(2021)]%
        {bi2021deep}
\bibfield{author}{\bibinfo{person}{Sai Bi}, \bibinfo{person}{Stephen Lombardi},
  \bibinfo{person}{Shunsuke Saito}, \bibinfo{person}{Tomas Simon},
  \bibinfo{person}{Shih-En Wei}, \bibinfo{person}{Kevyn Mcphail},
  \bibinfo{person}{Ravi Ramamoorthi}, \bibinfo{person}{Yaser Sheikh}, {and}
  \bibinfo{person}{Jason Saragih}.} \bibinfo{year}{2021}\natexlab{}.
\newblock \showarticletitle{Deep relightable appearance models for animatable
  faces}.
\newblock \bibinfo{journal}{\emph{ACM Transactions on Graphics (TOG)}}
  \bibinfo{volume}{40}, \bibinfo{number}{4} (\bibinfo{year}{2021}),
  \bibinfo{pages}{1--15}.
\newblock


\bibitem[Blanz and Vetter(1999)]%
        {blanz1999morphable}
\bibfield{author}{\bibinfo{person}{Volker Blanz} {and} \bibinfo{person}{Thomas
  Vetter}.} \bibinfo{year}{1999}\natexlab{}.
\newblock \showarticletitle{A morphable model for the synthesis of 3D faces}.
  In \bibinfo{booktitle}{\emph{Proceedings of the 26th annual conference on
  Computer graphics and interactive techniques}}. \bibinfo{pages}{187--194}.
\newblock


\bibitem[Burkov et~al\mbox{.}(2020)]%
        {burkov2020neural}
\bibfield{author}{\bibinfo{person}{Egor Burkov}, \bibinfo{person}{Igor
  Pasechnik}, \bibinfo{person}{Artur Grigorev}, {and} \bibinfo{person}{Victor
  Lempitsky}.} \bibinfo{year}{2020}\natexlab{}.
\newblock \showarticletitle{Neural head reenactment with latent pose
  descriptors}. In \bibinfo{booktitle}{\emph{Proceedings of the IEEE/CVF
  Conference on Computer Vision and Pattern Recognition}}.
  \bibinfo{pages}{13786--13795}.
\newblock


\bibitem[Cao et~al\mbox{.}(2021)]%
        {cao2021real}
\bibfield{author}{\bibinfo{person}{Chen Cao}, \bibinfo{person}{Vasu Agrawal},
  \bibinfo{person}{Fernando De~La~Torre}, \bibinfo{person}{Lele Chen},
  \bibinfo{person}{Jason Saragih}, \bibinfo{person}{Tomas Simon}, {and}
  \bibinfo{person}{Yaser Sheikh}.} \bibinfo{year}{2021}\natexlab{}.
\newblock \showarticletitle{Real-time 3D neural facial animation from binocular
  video}.
\newblock \bibinfo{journal}{\emph{ACM Transactions on Graphics (TOG)}}
  \bibinfo{volume}{40}, \bibinfo{number}{4} (\bibinfo{year}{2021}),
  \bibinfo{pages}{1--17}.
\newblock


\bibitem[Cao et~al\mbox{.}(2015)]%
        {cao2015real}
\bibfield{author}{\bibinfo{person}{Chen Cao}, \bibinfo{person}{Derek Bradley},
  \bibinfo{person}{Kun Zhou}, {and} \bibinfo{person}{Thabo Beeler}.}
  \bibinfo{year}{2015}\natexlab{}.
\newblock \showarticletitle{Real-time high-fidelity facial performance
  capture}.
\newblock \bibinfo{journal}{\emph{ACM Transactions on Graphics (ToG)}}
  \bibinfo{volume}{34}, \bibinfo{number}{4} (\bibinfo{year}{2015}),
  \bibinfo{pages}{1--9}.
\newblock


\bibitem[Cao et~al\mbox{.}(2022)]%
        {cao2022authentic}
\bibfield{author}{\bibinfo{person}{Chen Cao}, \bibinfo{person}{Tomas Simon},
  \bibinfo{person}{Jin~Kyu Kim}, \bibinfo{person}{Gabe Schwartz},
  \bibinfo{person}{Michael Zollhoefer}, \bibinfo{person}{Shun-Suke Saito},
  \bibinfo{person}{Stephen Lombardi}, \bibinfo{person}{Shih-En Wei},
  \bibinfo{person}{Danielle Belko}, \bibinfo{person}{Shoou-I Yu},
  {et~al\mbox{.}}} \bibinfo{year}{2022}\natexlab{}.
\newblock \showarticletitle{Authentic volumetric avatars from a phone scan}.
\newblock \bibinfo{journal}{\emph{ACM Transactions on Graphics (TOG)}}
  \bibinfo{volume}{41}, \bibinfo{number}{4} (\bibinfo{year}{2022}),
  \bibinfo{pages}{1--19}.
\newblock


\bibitem[Cao et~al\mbox{.}(2013)]%
        {cao20133d}
\bibfield{author}{\bibinfo{person}{Chen Cao}, \bibinfo{person}{Yanlin Weng},
  \bibinfo{person}{Stephen Lin}, {and} \bibinfo{person}{Kun Zhou}.}
  \bibinfo{year}{2013}\natexlab{}.
\newblock \showarticletitle{3D shape regression for real-time facial
  animation}.
\newblock \bibinfo{journal}{\emph{ACM Transactions on Graphics (TOG)}}
  \bibinfo{volume}{32}, \bibinfo{number}{4} (\bibinfo{year}{2013}),
  \bibinfo{pages}{1--10}.
\newblock


\bibitem[Cao et~al\mbox{.}(2014)]%
        {Cao2014Facewarehosue}
\bibfield{author}{\bibinfo{person}{Chen Cao}, \bibinfo{person}{Yanlin Weng},
  \bibinfo{person}{Shun Zhou}, \bibinfo{person}{Yiying Tong}, {and}
  \bibinfo{person}{Kun Zhou}.} \bibinfo{year}{2014}\natexlab{}.
\newblock \showarticletitle{FaceWarehouse: A 3D Facial Expression Database for
  Visual Computing}.
\newblock \bibinfo{journal}{\emph{IEEE Transactions on Visualization and
  Computer Graphics}} \bibinfo{volume}{20}, \bibinfo{number}{3}
  (\bibinfo{year}{2014}), \bibinfo{pages}{413--425}.
\newblock
\urldef\tempurl%
\url{https://doi.org/10.1109/TVCG.2013.249}
\showDOI{\tempurl}


\bibitem[Chandran et~al\mbox{.}(2020)]%
        {chandran2020semantic}
\bibfield{author}{\bibinfo{person}{Prashanth Chandran}, \bibinfo{person}{Derek
  Bradley}, \bibinfo{person}{Markus Gross}, {and} \bibinfo{person}{Thabo
  Beeler}.} \bibinfo{year}{2020}\natexlab{}.
\newblock \showarticletitle{Semantic deep face models}. In
  \bibinfo{booktitle}{\emph{2020 International Conference on 3D Vision (3DV)}}.
  IEEE, \bibinfo{pages}{345--354}.
\newblock


\bibitem[Chen et~al\mbox{.}(2019)]%
        {chen2019photo}
\bibfield{author}{\bibinfo{person}{Anpei Chen}, \bibinfo{person}{Zhang Chen},
  \bibinfo{person}{Guli Zhang}, \bibinfo{person}{Kenny Mitchell}, {and}
  \bibinfo{person}{Jingyi Yu}.} \bibinfo{year}{2019}\natexlab{}.
\newblock \showarticletitle{Photo-realistic facial details synthesis from
  single image}. In \bibinfo{booktitle}{\emph{Proceedings of the IEEE/CVF
  International Conference on Computer Vision}}. \bibinfo{pages}{9429--9439}.
\newblock


\bibitem[Chen et~al\mbox{.}(2021)]%
        {chen2021high}
\bibfield{author}{\bibinfo{person}{Lele Chen}, \bibinfo{person}{Chen Cao},
  \bibinfo{person}{Fernando De~la Torre}, \bibinfo{person}{Jason Saragih},
  \bibinfo{person}{Chenliang Xu}, {and} \bibinfo{person}{Yaser Sheikh}.}
  \bibinfo{year}{2021}\natexlab{}.
\newblock \showarticletitle{High-fidelity Face Tracking for AR/VR via Deep
  Lighting Adaptation}. In \bibinfo{booktitle}{\emph{Proceedings of the
  IEEE/CVF Conference on Computer Vision and Pattern Recognition}}.
  \bibinfo{pages}{13059--13069}.
\newblock


\bibitem[Cheng et~al\mbox{.}(2018)]%
        {Cheng2018}
\bibfield{author}{\bibinfo{person}{Shiyang Cheng}, \bibinfo{person}{Irene
  Kotsia}, \bibinfo{person}{Maja Pantic}, {and} \bibinfo{person}{Stefanos
  Zafeiriou}.} \bibinfo{year}{2018}\natexlab{}.
\newblock \showarticletitle{{4DFAB: A Large Scale 4D Database for Facial
  Expression Analysis and Biometric Applications}}. In
  \bibinfo{booktitle}{\emph{Proceedings of the IEEE Computer Society Conference
  on Computer Vision and Pattern Recognition}}. \bibinfo{pages}{5117--5126}.
\newblock
\showISBNx{9781538664209}
\showISSN{10636919}
\urldef\tempurl%
\url{https://doi.org/10.1109/CVPR.2018.00537}
\showDOI{\tempurl}


\bibitem[Cosker et~al\mbox{.}(2011)]%
        {Cosker2011}
\bibfield{author}{\bibinfo{person}{Darren Cosker}, \bibinfo{person}{Eva
  Krumhuber}, {and} \bibinfo{person}{Adrian Hilton}.}
  \bibinfo{year}{2011}\natexlab{}.
\newblock \showarticletitle{{A FACS valid 3D dynamic action unit database with
  applications to 3D dynamic morphable facial modeling}}. In
  \bibinfo{booktitle}{\emph{Proceedings of the IEEE International Conference on
  Computer Vision}}. \bibinfo{pages}{2296--2303}.
\newblock
\showISBNx{9781457711015}
\urldef\tempurl%
\url{https://doi.org/10.1109/ICCV.2011.6126510}
\showDOI{\tempurl}


\bibitem[Debevec et~al\mbox{.}(2000)]%
        {debevec2000acquiring}
\bibfield{author}{\bibinfo{person}{Paul Debevec}, \bibinfo{person}{Tim
  Hawkins}, \bibinfo{person}{Chris Tchou}, \bibinfo{person}{Haarm-Pieter
  Duiker}, \bibinfo{person}{Westley Sarokin}, {and} \bibinfo{person}{Mark
  Sagar}.} \bibinfo{year}{2000}\natexlab{}.
\newblock \showarticletitle{Acquiring the reflectance field of a human face}.
  In \bibinfo{booktitle}{\emph{Proceedings of the 27th annual conference on
  Computer graphics and interactive techniques}}. \bibinfo{pages}{145--156}.
\newblock


\bibitem[Egger et~al\mbox{.}(2020)]%
        {egger20203dmm}
\bibfield{author}{\bibinfo{person}{Bernhard Egger}, \bibinfo{person}{William~AP
  Smith}, \bibinfo{person}{Ayush Tewari}, \bibinfo{person}{Stefanie Wuhrer},
  \bibinfo{person}{Michael Zollhoefer}, \bibinfo{person}{Thabo Beeler},
  \bibinfo{person}{Florian Bernard}, \bibinfo{person}{Timo Bolkart},
  \bibinfo{person}{Adam Kortylewski}, \bibinfo{person}{Sami Romdhani},
  {et~al\mbox{.}}} \bibinfo{year}{2020}\natexlab{}.
\newblock \showarticletitle{3d morphable face models—past, present, and
  future}.
\newblock \bibinfo{journal}{\emph{ACM Transactions on Graphics (TOG)}}
  \bibinfo{volume}{39}, \bibinfo{number}{5} (\bibinfo{year}{2020}),
  \bibinfo{pages}{1--38}.
\newblock


\bibitem[Ekman and Friesen(1978)]%
        {ekman1978facial}
\bibfield{author}{\bibinfo{person}{Paul Ekman} {and} \bibinfo{person}{Wallace~V
  Friesen}.} \bibinfo{year}{1978}\natexlab{}.
\newblock \showarticletitle{Facial action coding system}.
\newblock \bibinfo{journal}{\emph{Environmental Psychology \& Nonverbal
  Behavior}} (\bibinfo{year}{1978}).
\newblock


\bibitem[Feng et~al\mbox{.}(2021)]%
        {DECA:Siggraph2021}
\bibfield{author}{\bibinfo{person}{Yao Feng}, \bibinfo{person}{Haiwen Feng},
  \bibinfo{person}{Michael~J. Black}, {and} \bibinfo{person}{Timo Bolkart}.}
  \bibinfo{year}{2021}\natexlab{}.
\newblock \showarticletitle{Learning an Animatable Detailed {3D} Face Model
  from In-The-Wild Images}.
\newblock \bibinfo{journal}{\emph{ACM Transactions on Graphics, (Proc.
  SIGGRAPH)}} \bibinfo{volume}{40}, \bibinfo{number}{8}.
\newblock
\urldef\tempurl%
\url{https://doi.org/10.1145/3450626.3459936}
\showURL{%
\tempurl}


\bibitem[Feng et~al\mbox{.}(2018)]%
        {feng2018joint}
\bibfield{author}{\bibinfo{person}{Yao Feng}, \bibinfo{person}{Fan Wu},
  \bibinfo{person}{Xiaohu Shao}, \bibinfo{person}{Yanfeng Wang}, {and}
  \bibinfo{person}{Xi Zhou}.} \bibinfo{year}{2018}\natexlab{}.
\newblock \showarticletitle{Joint 3d face reconstruction and dense alignment
  with position map regression network}. In
  \bibinfo{booktitle}{\emph{Proceedings of the European Conference on Computer
  Vision (ECCV)}}. \bibinfo{pages}{534--551}.
\newblock


\bibitem[Fyffe and Debevec(2015)]%
        {fyffe2015single}
\bibfield{author}{\bibinfo{person}{Graham Fyffe} {and} \bibinfo{person}{Paul
  Debevec}.} \bibinfo{year}{2015}\natexlab{}.
\newblock \showarticletitle{Single-shot reflectance measurement from polarized
  color gradient illumination}. In \bibinfo{booktitle}{\emph{2015 IEEE
  International Conference on Computational Photography (ICCP)}}. IEEE,
  \bibinfo{pages}{1--10}.
\newblock


\bibitem[Fyffe et~al\mbox{.}(2011)]%
        {fyffe2011comprehensive}
\bibfield{author}{\bibinfo{person}{Graham Fyffe}, \bibinfo{person}{Tim
  Hawkins}, \bibinfo{person}{Chris Watts}, \bibinfo{person}{Wan-Chun Ma}, {and}
  \bibinfo{person}{Paul Debevec}.} \bibinfo{year}{2011}\natexlab{}.
\newblock \showarticletitle{Comprehensive facial performance capture}. In
  \bibinfo{booktitle}{\emph{Computer Graphics Forum}},
  Vol.~\bibinfo{volume}{30}. Wiley Online Library, \bibinfo{pages}{425--434}.
\newblock


\bibitem[Fyffe et~al\mbox{.}(2014)]%
        {fyffe2014driving}
\bibfield{author}{\bibinfo{person}{Graham Fyffe}, \bibinfo{person}{Andrew
  Jones}, \bibinfo{person}{Oleg Alexander}, \bibinfo{person}{Ryosuke Ichikari},
  {and} \bibinfo{person}{Paul Debevec}.} \bibinfo{year}{2014}\natexlab{}.
\newblock \bibinfo{booktitle}{\emph{Driving high-resolution facial scans with
  video performance capture}}.
\newblock \bibinfo{type}{{T}echnical {R}eport}.
  \bibinfo{institution}{UNIVERSITY OF SOUTHERN CALIFORNIA LOS ANGELES}.
\newblock


\bibitem[Gafni et~al\mbox{.}(2021)]%
        {gafni2021dynamic}
\bibfield{author}{\bibinfo{person}{Guy Gafni}, \bibinfo{person}{Justus Thies},
  \bibinfo{person}{Michael Zollhofer}, {and} \bibinfo{person}{Matthias
  Nie{\ss}ner}.} \bibinfo{year}{2021}\natexlab{}.
\newblock \showarticletitle{Dynamic neural radiance fields for monocular 4d
  facial avatar reconstruction}. In \bibinfo{booktitle}{\emph{Proceedings of
  the IEEE/CVF Conference on Computer Vision and Pattern Recognition}}.
  \bibinfo{pages}{8649--8658}.
\newblock


\bibitem[Ganin and Lempitsky(2015)]%
        {pmlr-v37-ganin15}
\bibfield{author}{\bibinfo{person}{Yaroslav Ganin} {and}
  \bibinfo{person}{Victor Lempitsky}.} \bibinfo{year}{2015}\natexlab{}.
\newblock \showarticletitle{Unsupervised Domain Adaptation by Backpropagation}.
  In \bibinfo{booktitle}{\emph{Proceedings of the 32nd International Conference
  on Machine Learning}} \emph{(\bibinfo{series}{Proceedings of Machine Learning
  Research}, Vol.~\bibinfo{volume}{37})},
  \bibfield{editor}{\bibinfo{person}{Francis Bach} {and} \bibinfo{person}{David
  Blei}} (Eds.). \bibinfo{publisher}{PMLR}, \bibinfo{address}{Lille, France},
  \bibinfo{pages}{1180--1189}.
\newblock
\urldef\tempurl%
\url{https://proceedings.mlr.press/v37/ganin15.html}
\showURL{%
\tempurl}


\bibitem[Gecer et~al\mbox{.}(2019)]%
        {gecer2019ganfit}
\bibfield{author}{\bibinfo{person}{Baris Gecer}, \bibinfo{person}{Stylianos
  Ploumpis}, \bibinfo{person}{Irene Kotsia}, {and} \bibinfo{person}{Stefanos
  Zafeiriou}.} \bibinfo{year}{2019}\natexlab{}.
\newblock \showarticletitle{Ganfit: Generative adversarial network fitting for
  high fidelity 3d face reconstruction}. In
  \bibinfo{booktitle}{\emph{Proceedings of the IEEE/CVF Conference on Computer
  Vision and Pattern Recognition}}. \bibinfo{pages}{1155--1164}.
\newblock


\bibitem[Ghosh et~al\mbox{.}(2011)]%
        {ghosh2011multiview}
\bibfield{author}{\bibinfo{person}{Abhijeet Ghosh}, \bibinfo{person}{Graham
  Fyffe}, \bibinfo{person}{Borom Tunwattanapong}, \bibinfo{person}{Jay Busch},
  \bibinfo{person}{Xueming Yu}, {and} \bibinfo{person}{Paul Debevec}.}
  \bibinfo{year}{2011}\natexlab{}.
\newblock \showarticletitle{Multiview face capture using polarized spherical
  gradient illumination}. In \bibinfo{booktitle}{\emph{Proceedings of the 2011
  SIGGRAPH Asia Conference}}. \bibinfo{pages}{1--10}.
\newblock


\bibitem[Gotardo et~al\mbox{.}(2018)]%
        {gotardo2018practical}
\bibfield{author}{\bibinfo{person}{Paulo Gotardo},
  \bibinfo{person}{J{\'e}r{\'e}my Riviere}, \bibinfo{person}{Derek Bradley},
  \bibinfo{person}{Abhijeet Ghosh}, {and} \bibinfo{person}{Thabo Beeler}.}
  \bibinfo{year}{2018}\natexlab{}.
\newblock \showarticletitle{Practical dynamic facial appearance modeling and
  acquisition}.
\newblock  (\bibinfo{year}{2018}).
\newblock


\bibitem[Guo et~al\mbox{.}(2020)]%
        {guo2020towards}
\bibfield{author}{\bibinfo{person}{Jianzhu Guo}, \bibinfo{person}{Xiangyu Zhu},
  \bibinfo{person}{Yang Yang}, \bibinfo{person}{Fan Yang},
  \bibinfo{person}{Zhen Lei}, {and} \bibinfo{person}{Stan~Z Li}.}
  \bibinfo{year}{2020}\natexlab{}.
\newblock \showarticletitle{Towards Fast, Accurate and Stable 3D Dense Face
  Alignment}. In \bibinfo{booktitle}{\emph{Proceedings of the European
  Conference on Computer Vision (ECCV)}}.
\newblock


\bibitem[Guo et~al\mbox{.}(2019)]%
        {guo2019relightables}
\bibfield{author}{\bibinfo{person}{Kaiwen Guo}, \bibinfo{person}{Peter
  Lincoln}, \bibinfo{person}{Philip Davidson}, \bibinfo{person}{Jay Busch},
  \bibinfo{person}{Xueming Yu}, \bibinfo{person}{Matt Whalen},
  \bibinfo{person}{Geoff Harvey}, \bibinfo{person}{Sergio Orts-Escolano},
  \bibinfo{person}{Rohit Pandey}, \bibinfo{person}{Jason Dourgarian},
  {et~al\mbox{.}}} \bibinfo{year}{2019}\natexlab{}.
\newblock \showarticletitle{The relightables: Volumetric performance capture of
  humans with realistic relighting}.
\newblock \bibinfo{journal}{\emph{ACM Transactions on Graphics (TOG)}}
  \bibinfo{volume}{38}, \bibinfo{number}{6} (\bibinfo{year}{2019}),
  \bibinfo{pages}{1--19}.
\newblock


\bibitem[Guo et~al\mbox{.}(2018)]%
        {guo2018cnn}
\bibfield{author}{\bibinfo{person}{Yudong Guo}, \bibinfo{person}{Jianfei Cai},
  \bibinfo{person}{Boyi Jiang}, \bibinfo{person}{Jianmin Zheng},
  {et~al\mbox{.}}} \bibinfo{year}{2018}\natexlab{}.
\newblock \showarticletitle{Cnn-based real-time dense face reconstruction with
  inverse-rendered photo-realistic face images}.
\newblock \bibinfo{journal}{\emph{IEEE transactions on pattern analysis and
  machine intelligence}} \bibinfo{volume}{41}, \bibinfo{number}{6}
  (\bibinfo{year}{2018}), \bibinfo{pages}{1294--1307}.
\newblock


\bibitem[Higgins et~al\mbox{.}(2016)]%
        {higgins2016beta}
\bibfield{author}{\bibinfo{person}{Irina Higgins}, \bibinfo{person}{Loic
  Matthey}, \bibinfo{person}{Arka Pal}, \bibinfo{person}{Christopher Burgess},
  \bibinfo{person}{Xavier Glorot}, \bibinfo{person}{Matthew Botvinick},
  \bibinfo{person}{Shakir Mohamed}, {and} \bibinfo{person}{Alexander
  Lerchner}.} \bibinfo{year}{2016}\natexlab{}.
\newblock \showarticletitle{beta-vae: Learning basic visual concepts with a
  constrained variational framework}.
\newblock  (\bibinfo{year}{2016}).
\newblock


\bibitem[Huang et~al\mbox{.}(2018)]%
        {huang2018multimodal}
\bibfield{author}{\bibinfo{person}{Xun Huang}, \bibinfo{person}{Ming-Yu Liu},
  \bibinfo{person}{Serge Belongie}, {and} \bibinfo{person}{Jan Kautz}.}
  \bibinfo{year}{2018}\natexlab{}.
\newblock \showarticletitle{Multimodal unsupervised image-to-image
  translation}. In \bibinfo{booktitle}{\emph{Proceedings of the European
  conference on computer vision (ECCV)}}. \bibinfo{pages}{172--189}.
\newblock


\bibitem[Huynh et~al\mbox{.}(2018)]%
        {huynh2018mesoscopic}
\bibfield{author}{\bibinfo{person}{Loc Huynh}, \bibinfo{person}{Weikai Chen},
  \bibinfo{person}{Shunsuke Saito}, \bibinfo{person}{Jun Xing},
  \bibinfo{person}{Koki Nagano}, \bibinfo{person}{Andrew Jones},
  \bibinfo{person}{Paul Debevec}, {and} \bibinfo{person}{Hao Li}.}
  \bibinfo{year}{2018}\natexlab{}.
\newblock \showarticletitle{Mesoscopic facial geometry inference using deep
  neural networks}. In \bibinfo{booktitle}{\emph{Proceedings of the IEEE
  Conference on Computer Vision and Pattern Recognition}}.
  \bibinfo{pages}{8407--8416}.
\newblock


\bibitem[Joo et~al\mbox{.}(2017)]%
        {Joo_2017_TPAMI}
\bibfield{author}{\bibinfo{person}{Hanbyul Joo}, \bibinfo{person}{Tomas Simon},
  \bibinfo{person}{Xulong Li}, \bibinfo{person}{Hao Liu}, \bibinfo{person}{Lei
  Tan}, \bibinfo{person}{Lin Gui}, \bibinfo{person}{Sean Banerjee},
  \bibinfo{person}{Timothy~Scott Godisart}, \bibinfo{person}{Bart Nabbe},
  \bibinfo{person}{Iain Matthews}, \bibinfo{person}{Takeo Kanade},
  \bibinfo{person}{Shohei Nobuhara}, {and} \bibinfo{person}{Yaser Sheikh}.}
  \bibinfo{year}{2017}\natexlab{}.
\newblock \showarticletitle{Panoptic Studio: A Massively Multiview System for
  Social Interaction Capture}.
\newblock \bibinfo{journal}{\emph{IEEE Transactions on Pattern Analysis and
  Machine Intelligence}} (\bibinfo{year}{2017}).
\newblock


\bibitem[King(2009)]%
        {king2009dlib}
\bibfield{author}{\bibinfo{person}{Davis~E King}.}
  \bibinfo{year}{2009}\natexlab{}.
\newblock \showarticletitle{Dlib-ml: A machine learning toolkit}.
\newblock \bibinfo{journal}{\emph{The Journal of Machine Learning Research}}
  \bibinfo{volume}{10} (\bibinfo{year}{2009}), \bibinfo{pages}{1755--1758}.
\newblock


\bibitem[Kingma and Welling(2014)]%
        {Kingma2014}
\bibfield{author}{\bibinfo{person}{Diederik~P. Kingma} {and}
  \bibinfo{person}{Max Welling}.} \bibinfo{year}{2014}\natexlab{}.
\newblock \showarticletitle{{Auto-Encoding Variational Bayes}}. In
  \bibinfo{booktitle}{\emph{2nd International Conference on Learning
  Representations, {ICLR} 2014, Banff, AB, Canada, April 14-16, 2014,
  Conference Track Proceedings}}.
\newblock


\bibitem[Klaudiny et~al\mbox{.}(2017)]%
        {klaudiny2017real}
\bibfield{author}{\bibinfo{person}{Martin Klaudiny}, \bibinfo{person}{Steven
  McDonagh}, \bibinfo{person}{Derek Bradley}, \bibinfo{person}{Thabo Beeler},
  {and} \bibinfo{person}{Kenny Mitchell}.} \bibinfo{year}{2017}\natexlab{}.
\newblock \showarticletitle{Real-Time Multi-View Facial Capture with Synthetic
  Training}. In \bibinfo{booktitle}{\emph{Computer Graphics Forum}},
  Vol.~\bibinfo{volume}{36}. Wiley Online Library, \bibinfo{pages}{325--336}.
\newblock


\bibitem[Laine et~al\mbox{.}(2017)]%
        {laine2017production}
\bibfield{author}{\bibinfo{person}{Samuli Laine}, \bibinfo{person}{Tero
  Karras}, \bibinfo{person}{Timo Aila}, \bibinfo{person}{Antti Herva},
  \bibinfo{person}{Shunsuke Saito}, \bibinfo{person}{Ronald Yu},
  \bibinfo{person}{Hao Li}, {and} \bibinfo{person}{Jaakko Lehtinen}.}
  \bibinfo{year}{2017}\natexlab{}.
\newblock \showarticletitle{Production-level facial performance capture using
  deep convolutional neural networks}. In \bibinfo{booktitle}{\emph{Proceedings
  of the ACM SIGGRAPH/Eurographics symposium on computer animation}}.
  \bibinfo{pages}{1--10}.
\newblock


\bibitem[Lattas et~al\mbox{.}(2021)]%
        {lattas2021avatarme++}
\bibfield{author}{\bibinfo{person}{Alexandros Lattas},
  \bibinfo{person}{Stylianos Moschoglou}, \bibinfo{person}{Stylianos Ploumpis},
  \bibinfo{person}{Baris Gecer}, \bibinfo{person}{Abhijeet Ghosh}, {and}
  \bibinfo{person}{Stefanos~P Zafeiriou}.} \bibinfo{year}{2021}\natexlab{}.
\newblock \showarticletitle{AvatarMe++: Facial Shape and BRDF Inference with
  Photorealistic Rendering-Aware GANs}.
\newblock \bibinfo{journal}{\emph{IEEE Transactions on Pattern Analysis \&
  Machine Intelligence}} \bibinfo{number}{01} (\bibinfo{year}{2021}),
  \bibinfo{pages}{1--1}.
\newblock


\bibitem[LeGendre et~al\mbox{.}(2018)]%
        {legendre2018efficient}
\bibfield{author}{\bibinfo{person}{Chloe LeGendre}, \bibinfo{person}{Kalle
  Bladin}, \bibinfo{person}{Bipin Kishore}, \bibinfo{person}{Xinglei Ren},
  \bibinfo{person}{Xueming Yu}, {and} \bibinfo{person}{Paul Debevec}.}
  \bibinfo{year}{2018}\natexlab{}.
\newblock \showarticletitle{Efficient multispectral facial capture with
  monochrome cameras}. In \bibinfo{booktitle}{\emph{Color and Imaging
  Conference}}, Vol.~\bibinfo{volume}{2018}. Society for Imaging Science and
  Technology, \bibinfo{pages}{187--202}.
\newblock


\bibitem[Lewis et~al\mbox{.}(2014)]%
        {lewis2014practice}
\bibfield{author}{\bibinfo{person}{John~P Lewis}, \bibinfo{person}{Ken Anjyo},
  \bibinfo{person}{Taehyun Rhee}, \bibinfo{person}{Mengjie Zhang},
  \bibinfo{person}{Frederic~H Pighin}, {and} \bibinfo{person}{Zhigang Deng}.}
  \bibinfo{year}{2014}\natexlab{}.
\newblock \showarticletitle{Practice and Theory of Blendshape Facial Models.}
\newblock \bibinfo{journal}{\emph{Eurographics (State of the Art Reports)}}
  \bibinfo{volume}{1}, \bibinfo{number}{8} (\bibinfo{year}{2014}),
  \bibinfo{pages}{2}.
\newblock


\bibitem[Li et~al\mbox{.}(2020b)]%
        {li2020dynamic}
\bibfield{author}{\bibinfo{person}{Jiaman Li}, \bibinfo{person}{Zhengfei
  Kuang}, \bibinfo{person}{Yajie Zhao}, \bibinfo{person}{Mingming He},
  \bibinfo{person}{Karl Bladin}, {and} \bibinfo{person}{Hao Li}.}
  \bibinfo{year}{2020}\natexlab{b}.
\newblock \showarticletitle{Dynamic facial asset and rig generation from a
  single scan.}
\newblock \bibinfo{journal}{\emph{ACM Trans. Graph.}} \bibinfo{volume}{39},
  \bibinfo{number}{6} (\bibinfo{year}{2020}), \bibinfo{pages}{215--1}.
\newblock


\bibitem[Li et~al\mbox{.}(2020a)]%
        {li2020learning}
\bibfield{author}{\bibinfo{person}{Ruilong Li}, \bibinfo{person}{Karl Bladin},
  \bibinfo{person}{Yajie Zhao}, \bibinfo{person}{Chinmay Chinara},
  \bibinfo{person}{Owen Ingraham}, \bibinfo{person}{Pengda Xiang},
  \bibinfo{person}{Xinglei Ren}, \bibinfo{person}{Pratusha Prasad},
  \bibinfo{person}{Bipin Kishore}, \bibinfo{person}{Jun Xing}, {et~al\mbox{.}}}
  \bibinfo{year}{2020}\natexlab{a}.
\newblock \showarticletitle{Learning formation of physically-based face
  attributes}. In \bibinfo{booktitle}{\emph{Proceedings of the IEEE/CVF
  Conference on Computer Vision and Pattern Recognition}}.
  \bibinfo{pages}{3410--3419}.
\newblock


\bibitem[Li et~al\mbox{.}(2017)]%
        {FLAME:SiggraphAsia2017}
\bibfield{author}{\bibinfo{person}{Tianye Li}, \bibinfo{person}{Timo Bolkart},
  \bibinfo{person}{Michael.~J. Black}, \bibinfo{person}{Hao Li}, {and}
  \bibinfo{person}{Javier Romero}.} \bibinfo{year}{2017}\natexlab{}.
\newblock \showarticletitle{Learning a model of facial shape and expression
  from {4D} scans}.
\newblock \bibinfo{journal}{\emph{ACM Transactions on Graphics, (Proc. SIGGRAPH
  Asia)}} \bibinfo{volume}{36}, \bibinfo{number}{6} (\bibinfo{year}{2017}),
  \bibinfo{pages}{194:1--194:17}.
\newblock
\urldef\tempurl%
\url{https://doi.org/10.1145/3130800.3130813}
\showURL{%
\tempurl}


\bibitem[Li et~al\mbox{.}(2021)]%
        {li2021topologically}
\bibfield{author}{\bibinfo{person}{Tianye Li}, \bibinfo{person}{Shichen Liu},
  \bibinfo{person}{Timo Bolkart}, \bibinfo{person}{Jiayi Liu},
  \bibinfo{person}{Hao Li}, {and} \bibinfo{person}{Yajie Zhao}.}
  \bibinfo{year}{2021}\natexlab{}.
\newblock \showarticletitle{Topologically Consistent Multi-View Face Inference
  Using Volumetric Sampling}. In \bibinfo{booktitle}{\emph{Proceedings of the
  IEEE/CVF International Conference on Computer Vision}}.
  \bibinfo{pages}{3824--3834}.
\newblock


\bibitem[Lombardi et~al\mbox{.}(2018)]%
        {lombardi2018deep}
\bibfield{author}{\bibinfo{person}{Stephen Lombardi}, \bibinfo{person}{Jason
  Saragih}, \bibinfo{person}{Tomas Simon}, {and} \bibinfo{person}{Yaser
  Sheikh}.} \bibinfo{year}{2018}\natexlab{}.
\newblock \showarticletitle{Deep appearance models for face rendering}.
\newblock \bibinfo{journal}{\emph{ACM Transactions on Graphics (TOG)}}
  \bibinfo{volume}{37}, \bibinfo{number}{4} (\bibinfo{year}{2018}),
  \bibinfo{pages}{1--13}.
\newblock


\bibitem[Lombardi et~al\mbox{.}(2019)]%
        {Lombardi2019}
\bibfield{author}{\bibinfo{person}{Stephen Lombardi}, \bibinfo{person}{Tomas
  Simon}, \bibinfo{person}{Jason Saragih}, \bibinfo{person}{Gabriel Schwartz},
  \bibinfo{person}{Andreas Lehrmann}, {and} \bibinfo{person}{Yaser Sheikh}.}
  \bibinfo{year}{2019}\natexlab{}.
\newblock \showarticletitle{Neural Volumes: Learning Dynamic Renderable Volumes
  from Images}.
\newblock \bibinfo{journal}{\emph{ACM Trans. Graph.}} \bibinfo{volume}{38},
  \bibinfo{number}{4}, Article \bibinfo{articleno}{65} (\bibinfo{date}{July}
  \bibinfo{year}{2019}), \bibinfo{numpages}{14}~pages.
\newblock


\bibitem[Lombardi et~al\mbox{.}(2021)]%
        {Lombardi21}
\bibfield{author}{\bibinfo{person}{Stephen Lombardi}, \bibinfo{person}{Tomas
  Simon}, \bibinfo{person}{Gabriel Schwartz}, \bibinfo{person}{Michael
  Zollhoefer}, \bibinfo{person}{Yaser Sheikh}, {and} \bibinfo{person}{Jason
  Saragih}.} \bibinfo{year}{2021}\natexlab{}.
\newblock \showarticletitle{Mixture of Volumetric Primitives for Efficient
  Neural Rendering}.
\newblock \bibinfo{journal}{\emph{ACM Trans. Graph.}} \bibinfo{volume}{40},
  \bibinfo{number}{4}, Article \bibinfo{articleno}{59} (\bibinfo{date}{jul}
  \bibinfo{year}{2021}), \bibinfo{numpages}{13}~pages.
\newblock
\showISSN{0730-0301}
\urldef\tempurl%
\url{https://doi.org/10.1145/3450626.3459863}
\showDOI{\tempurl}


\bibitem[Loper et~al\mbox{.}(2015)]%
        {SMPL:2015}
\bibfield{author}{\bibinfo{person}{Matthew Loper}, \bibinfo{person}{Naureen
  Mahmood}, \bibinfo{person}{Javier Romero}, \bibinfo{person}{Gerard
  Pons-Moll}, {and} \bibinfo{person}{Michael~J. Black}.}
  \bibinfo{year}{2015}\natexlab{}.
\newblock \showarticletitle{{SMPL}: A Skinned Multi-Person Linear Model}.
\newblock \bibinfo{journal}{\emph{ACM Trans. Graphics (Proc. SIGGRAPH Asia)}}
  \bibinfo{volume}{34}, \bibinfo{number}{6} (\bibinfo{date}{Oct.}
  \bibinfo{year}{2015}), \bibinfo{pages}{248:1--248:16}.
\newblock


\bibitem[Luo et~al\mbox{.}(2021)]%
        {luo2021convolutional}
\bibfield{author}{\bibinfo{person}{H. Luo}, \bibinfo{person}{A. Chen},
  \bibinfo{person}{Q. Zhang}, \bibinfo{person}{B. Pang}, \bibinfo{person}{M.
  Wu}, \bibinfo{person}{L. Xu}, {and} \bibinfo{person}{J. Yu}.}
  \bibinfo{year}{2021}\natexlab{}.
\newblock \showarticletitle{Convolutional Neural Opacity Radiance Fields}. In
  \bibinfo{booktitle}{\emph{2021 IEEE International Conference on Computational
  Photography (ICCP)}}. \bibinfo{publisher}{IEEE Computer Society},
  \bibinfo{address}{Los Alamitos, CA, USA}, \bibinfo{pages}{1--12}.
\newblock
\urldef\tempurl%
\url{https://doi.org/10.1109/ICCP51581.2021.9466273}
\showDOI{\tempurl}


\bibitem[Ma et~al\mbox{.}(2021)]%
        {ma2021pixel}
\bibfield{author}{\bibinfo{person}{Shugao Ma}, \bibinfo{person}{Tomas Simon},
  \bibinfo{person}{Jason Saragih}, \bibinfo{person}{Dawei Wang},
  \bibinfo{person}{Yuecheng Li}, \bibinfo{person}{Fernando De~La~Torre}, {and}
  \bibinfo{person}{Yaser Sheikh}.} \bibinfo{year}{2021}\natexlab{}.
\newblock \showarticletitle{Pixel Codec Avatars}. In
  \bibinfo{booktitle}{\emph{Proceedings of the IEEE/CVF Conference on Computer
  Vision and Pattern Recognition}}. \bibinfo{pages}{64--73}.
\newblock


\bibitem[Ma et~al\mbox{.}(2007)]%
        {ma2007rapid}
\bibfield{author}{\bibinfo{person}{Wan-Chun Ma}, \bibinfo{person}{Tim Hawkins},
  \bibinfo{person}{Pieter Peers}, \bibinfo{person}{Charles-Felix Chabert},
  \bibinfo{person}{Malte Weiss}, \bibinfo{person}{Paul~E Debevec},
  {et~al\mbox{.}}} \bibinfo{year}{2007}\natexlab{}.
\newblock \showarticletitle{Rapid Acquisition of Specular and Diffuse Normal
  Maps from Polarized Spherical Gradient Illumination.}
\newblock \bibinfo{journal}{\emph{Rendering Techniques}}
  \bibinfo{volume}{2007}, \bibinfo{number}{9} (\bibinfo{year}{2007}),
  \bibinfo{pages}{10}.
\newblock


\bibitem[Mildenhall et~al\mbox{.}(2020)]%
        {mildenhall2020nerf}
\bibfield{author}{\bibinfo{person}{Ben Mildenhall}, \bibinfo{person}{Pratul~P
  Srinivasan}, \bibinfo{person}{Matthew Tancik}, \bibinfo{person}{Jonathan~T
  Barron}, \bibinfo{person}{Ravi Ramamoorthi}, {and} \bibinfo{person}{Ren Ng}.}
  \bibinfo{year}{2020}\natexlab{}.
\newblock \showarticletitle{Nerf: Representing scenes as neural radiance fields
  for view synthesis}. In \bibinfo{booktitle}{\emph{European conference on
  computer vision}}. Springer, \bibinfo{pages}{405--421}.
\newblock


\bibitem[milesial(2022)]%
        {Pytorch-UNet}
\bibfield{author}{\bibinfo{person}{milesial}.} \bibinfo{year}{2022}\natexlab{}.
\newblock \bibinfo{title}{Pytorch-UNet}.
\newblock
  \bibinfo{howpublished}{\url{https://github.com/milesial/Pytorch-UNet}}.
\newblock


\bibitem[Morales et~al\mbox{.}(2021)]%
        {morales2021survey}
\bibfield{author}{\bibinfo{person}{Araceli Morales}, \bibinfo{person}{Gemma
  Piella}, {and} \bibinfo{person}{Federico~M Sukno}.}
  \bibinfo{year}{2021}\natexlab{}.
\newblock \showarticletitle{Survey on 3D face reconstruction from uncalibrated
  images}.
\newblock \bibinfo{journal}{\emph{Computer Science Review}}
  \bibinfo{volume}{40} (\bibinfo{year}{2021}), \bibinfo{pages}{100400}.
\newblock


\bibitem[Moser et~al\mbox{.}(2021)]%
        {moser2021semi}
\bibfield{author}{\bibinfo{person}{Lucio Moser}, \bibinfo{person}{Chinyu
  Chien}, \bibinfo{person}{Mark Williams}, \bibinfo{person}{Jose Serra},
  \bibinfo{person}{Darren Hendler}, {and} \bibinfo{person}{Doug Roble}.}
  \bibinfo{year}{2021}\natexlab{}.
\newblock \showarticletitle{Semi-supervised video-driven facial animation
  transfer for production}.
\newblock \bibinfo{journal}{\emph{ACM Transactions on Graphics (TOG)}}
  \bibinfo{volume}{40}, \bibinfo{number}{6} (\bibinfo{year}{2021}),
  \bibinfo{pages}{1--18}.
\newblock


\bibitem[Nirkin et~al\mbox{.}(2019)]%
        {nirkin2019fsgan}
\bibfield{author}{\bibinfo{person}{Yuval Nirkin}, \bibinfo{person}{Yosi
  Keller}, {and} \bibinfo{person}{Tal Hassner}.}
  \bibinfo{year}{2019}\natexlab{}.
\newblock \showarticletitle{Fsgan: Subject agnostic face swapping and
  reenactment}. In \bibinfo{booktitle}{\emph{Proceedings of the IEEE/CVF
  international conference on computer vision}}. \bibinfo{pages}{7184--7193}.
\newblock


\bibitem[Olszewski et~al\mbox{.}(2016)]%
        {olszewski2016high}
\bibfield{author}{\bibinfo{person}{Kyle Olszewski}, \bibinfo{person}{Joseph~J
  Lim}, \bibinfo{person}{Shunsuke Saito}, {and} \bibinfo{person}{Hao Li}.}
  \bibinfo{year}{2016}\natexlab{}.
\newblock \showarticletitle{High-fidelity facial and speech animation for VR
  HMDs}.
\newblock \bibinfo{journal}{\emph{ACM Transactions on Graphics (TOG)}}
  \bibinfo{volume}{35}, \bibinfo{number}{6} (\bibinfo{year}{2016}),
  \bibinfo{pages}{1--14}.
\newblock


\bibitem[Park et~al\mbox{.}(2020)]%
        {park2020swapping}
\bibfield{author}{\bibinfo{person}{Taesung Park}, \bibinfo{person}{Jun-Yan
  Zhu}, \bibinfo{person}{Oliver Wang}, \bibinfo{person}{Jingwan Lu},
  \bibinfo{person}{Eli Shechtman}, \bibinfo{person}{Alexei Efros}, {and}
  \bibinfo{person}{Richard Zhang}.} \bibinfo{year}{2020}\natexlab{}.
\newblock \showarticletitle{Swapping autoencoder for deep image manipulation}.
\newblock \bibinfo{journal}{\emph{Advances in Neural Information Processing
  Systems}}  \bibinfo{volume}{33} (\bibinfo{year}{2020}),
  \bibinfo{pages}{7198--7211}.
\newblock


\bibitem[Paysan et~al\mbox{.}(2009)]%
        {paysan20093d}
\bibfield{author}{\bibinfo{person}{Pascal Paysan}, \bibinfo{person}{Reinhard
  Knothe}, \bibinfo{person}{Brian Amberg}, \bibinfo{person}{Sami Romdhani},
  {and} \bibinfo{person}{Thomas Vetter}.} \bibinfo{year}{2009}\natexlab{}.
\newblock \showarticletitle{A 3D face model for pose and illumination invariant
  face recognition}. In \bibinfo{booktitle}{\emph{2009 sixth IEEE international
  conference on advanced video and signal based surveillance}}. Ieee,
  \bibinfo{pages}{296--301}.
\newblock


\bibitem[Raj et~al\mbox{.}(2021)]%
        {raj2021pixel}
\bibfield{author}{\bibinfo{person}{Amit Raj}, \bibinfo{person}{Michael
  Zollhofer}, \bibinfo{person}{Tomas Simon}, \bibinfo{person}{Jason Saragih},
  \bibinfo{person}{Shunsuke Saito}, \bibinfo{person}{James Hays}, {and}
  \bibinfo{person}{Stephen Lombardi}.} \bibinfo{year}{2021}\natexlab{}.
\newblock \showarticletitle{Pixel-Aligned Volumetric Avatars}. In
  \bibinfo{booktitle}{\emph{Proceedings of the IEEE/CVF Conference on Computer
  Vision and Pattern Recognition}}. \bibinfo{pages}{11733--11742}.
\newblock


\bibitem[Richardson et~al\mbox{.}(2016)]%
        {richardson20163d}
\bibfield{author}{\bibinfo{person}{Elad Richardson}, \bibinfo{person}{Matan
  Sela}, {and} \bibinfo{person}{Ron Kimmel}.} \bibinfo{year}{2016}\natexlab{}.
\newblock \showarticletitle{3D face reconstruction by learning from synthetic
  data}. In \bibinfo{booktitle}{\emph{2016 fourth international conference on
  3D vision (3DV)}}. IEEE, \bibinfo{pages}{460--469}.
\newblock


\bibitem[Riviere et~al\mbox{.}(2020)]%
        {riviere2020single}
\bibfield{author}{\bibinfo{person}{J{\'e}r{\'e}my Riviere},
  \bibinfo{person}{Paulo Gotardo}, \bibinfo{person}{Derek Bradley},
  \bibinfo{person}{Abhijeet Ghosh}, {and} \bibinfo{person}{Thabo Beeler}.}
  \bibinfo{year}{2020}\natexlab{}.
\newblock \showarticletitle{Single-shot high-quality facial geometry and skin
  appearance capture}.
\newblock  (\bibinfo{year}{2020}).
\newblock


\bibitem[Saito et~al\mbox{.}(2017)]%
        {saito2017photorealistic}
\bibfield{author}{\bibinfo{person}{Shunsuke Saito}, \bibinfo{person}{Lingyu
  Wei}, \bibinfo{person}{Liwen Hu}, \bibinfo{person}{Koki Nagano}, {and}
  \bibinfo{person}{Hao Li}.} \bibinfo{year}{2017}\natexlab{}.
\newblock \showarticletitle{Photorealistic facial texture inference using deep
  neural networks}. In \bibinfo{booktitle}{\emph{Proceedings of the IEEE
  Conference on Computer Vision and Pattern Recognition}}.
  \bibinfo{pages}{5144--5153}.
\newblock


\bibitem[Smith et~al\mbox{.}(2020)]%
        {smith2020morphable}
\bibfield{author}{\bibinfo{person}{William~AP Smith}, \bibinfo{person}{Alassane
  Seck}, \bibinfo{person}{Hannah Dee}, \bibinfo{person}{Bernard Tiddeman},
  \bibinfo{person}{Joshua~B Tenenbaum}, {and} \bibinfo{person}{Bernhard
  Egger}.} \bibinfo{year}{2020}\natexlab{}.
\newblock \showarticletitle{A morphable face albedo model}. In
  \bibinfo{booktitle}{\emph{Proceedings of the IEEE/CVF Conference on Computer
  Vision and Pattern Recognition}}. \bibinfo{pages}{5011--5020}.
\newblock


\bibitem[Stratou et~al\mbox{.}(2011)]%
        {3drfe}
\bibfield{author}{\bibinfo{person}{Giota Stratou}, \bibinfo{person}{Abhijeet
  Ghosh}, \bibinfo{person}{Paul Debevec}, {and} \bibinfo{person}{Louis-Philippe
  Morency}.} \bibinfo{year}{2011}\natexlab{}.
\newblock \showarticletitle{Effect of illumination on automatic expression
  recognition: A novel 3D relightable facial database}. In
  \bibinfo{booktitle}{\emph{2011 IEEE International Conference on Automatic
  Face Gesture Recognition (FG)}}. \bibinfo{pages}{611--618}.
\newblock
\urldef\tempurl%
\url{https://doi.org/10.1109/FG.2011.5771467}
\showDOI{\tempurl}


\bibitem[Tuan~Tran et~al\mbox{.}(2017)]%
        {tuan2017regressing}
\bibfield{author}{\bibinfo{person}{Anh Tuan~Tran}, \bibinfo{person}{Tal
  Hassner}, \bibinfo{person}{Iacopo Masi}, {and} \bibinfo{person}{G{\'e}rard
  Medioni}.} \bibinfo{year}{2017}\natexlab{}.
\newblock \showarticletitle{Regressing robust and discriminative 3D morphable
  models with a very deep neural network}. In
  \bibinfo{booktitle}{\emph{Proceedings of the IEEE conference on computer
  vision and pattern recognition}}. \bibinfo{pages}{5163--5172}.
\newblock


\bibitem[{Unity Technologies}(2020)]%
        {Heretic}
\bibfield{author}{\bibinfo{person}{{Unity Technologies}}.}
  \bibinfo{year}{2020}\natexlab{}.
\newblock \bibinfo{title}{The Heretic: Digital Human}.
\newblock
  \bibinfo{howpublished}{\url{https://assetstore.unity.com/packages/essentials/tutorial-projects/the-heretic-digital-human-168620}}.
\newblock


\bibitem[Wei et~al\mbox{.}(2019)]%
        {VRface}
\bibfield{author}{\bibinfo{person}{Shih-En Wei}, \bibinfo{person}{Jason
  Saragih}, \bibinfo{person}{Tomas Simon}, \bibinfo{person}{Adam~W. Harley},
  \bibinfo{person}{Stephen Lombardi}, \bibinfo{person}{Michal Perdoch},
  \bibinfo{person}{Alexander Hypes}, \bibinfo{person}{Dawei Wang},
  \bibinfo{person}{Hernan Badino}, {and} \bibinfo{person}{Yaser Sheikh}.}
  \bibinfo{year}{2019}\natexlab{}.
\newblock \showarticletitle{VR Facial Animation via Multiview Image
  Translation}.
\newblock  \bibinfo{volume}{38}, \bibinfo{number}{4}, Article
  \bibinfo{articleno}{67} (\bibinfo{date}{jul} \bibinfo{year}{2019}),
  \bibinfo{numpages}{16}~pages.
\newblock
\showISSN{0730-0301}
\urldef\tempurl%
\url{https://doi.org/10.1145/3306346.3323030}
\showDOI{\tempurl}


\bibitem[Weise et~al\mbox{.}(2011)]%
        {weise2011realtime}
\bibfield{author}{\bibinfo{person}{Thibaut Weise}, \bibinfo{person}{Sofien
  Bouaziz}, \bibinfo{person}{Hao Li}, {and} \bibinfo{person}{Mark Pauly}.}
  \bibinfo{year}{2011}\natexlab{}.
\newblock \showarticletitle{Realtime performance-based facial animation}.
\newblock \bibinfo{journal}{\emph{ACM transactions on graphics (TOG)}}
  \bibinfo{volume}{30}, \bibinfo{number}{4} (\bibinfo{year}{2011}),
  \bibinfo{pages}{1--10}.
\newblock


\bibitem[Wu et~al\mbox{.}(2016)]%
        {wu2016anatomically}
\bibfield{author}{\bibinfo{person}{Chenglei Wu}, \bibinfo{person}{Derek
  Bradley}, \bibinfo{person}{Markus Gross}, {and} \bibinfo{person}{Thabo
  Beeler}.} \bibinfo{year}{2016}\natexlab{}.
\newblock \showarticletitle{An anatomically-constrained local deformation model
  for monocular face capture}.
\newblock \bibinfo{journal}{\emph{ACM transactions on graphics (TOG)}}
  \bibinfo{volume}{35}, \bibinfo{number}{4} (\bibinfo{year}{2016}),
  \bibinfo{pages}{1--12}.
\newblock


\bibitem[Yamaguchi et~al\mbox{.}(2018)]%
        {yamaguchi2018high}
\bibfield{author}{\bibinfo{person}{Shugo Yamaguchi}, \bibinfo{person}{Shunsuke
  Saito}, \bibinfo{person}{Koki Nagano}, \bibinfo{person}{Yajie Zhao},
  \bibinfo{person}{Weikai Chen}, \bibinfo{person}{Kyle Olszewski},
  \bibinfo{person}{Shigeo Morishima}, {and} \bibinfo{person}{Hao Li}.}
  \bibinfo{year}{2018}\natexlab{}.
\newblock \showarticletitle{High-fidelity facial reflectance and geometry
  inference from an unconstrained image}.
\newblock \bibinfo{journal}{\emph{ACM Transactions on Graphics (TOG)}}
  \bibinfo{volume}{37}, \bibinfo{number}{4} (\bibinfo{year}{2018}),
  \bibinfo{pages}{1--14}.
\newblock


\bibitem[Yang et~al\mbox{.}(2020)]%
        {yang2020facescape}
\bibfield{author}{\bibinfo{person}{Haotian Yang}, \bibinfo{person}{Hao Zhu},
  \bibinfo{person}{Yanru Wang}, \bibinfo{person}{Mingkai Huang},
  \bibinfo{person}{Qiu Shen}, \bibinfo{person}{Ruigang Yang}, {and}
  \bibinfo{person}{Xun Cao}.} \bibinfo{year}{2020}\natexlab{}.
\newblock \showarticletitle{FaceScape: a Large-scale High Quality 3D Face
  Dataset and Detailed Riggable 3D Face Prediction}. In
  \bibinfo{booktitle}{\emph{Proceedings of the IEEE Conference on Computer
  Vision and Pattern Recognition (CVPR)}}.
\newblock


\bibitem[Yin et~al\mbox{.}(2009)]%
        {yin2009bjut}
\bibfield{author}{\bibinfo{person}{Baocai Yin}, \bibinfo{person}{Yanfeng Sun},
  \bibinfo{person}{Chengzhang Wang}, {and} \bibinfo{person}{Yun Ge}.}
  \bibinfo{year}{2009}\natexlab{}.
\newblock \showarticletitle{BJUT-3D large scale 3D face database and
  information processing}.
\newblock \bibinfo{journal}{\emph{Journal of Computer Research and
  Development}} \bibinfo{volume}{46}, \bibinfo{number}{6}
  (\bibinfo{year}{2009}), \bibinfo{pages}{1009}.
\newblock


\bibitem[Yin et~al\mbox{.}(2008)]%
        {yin20084d}
\bibfield{author}{\bibinfo{person}{Lijun Yin}, \bibinfo{person}{Xiaochen Chen},
  \bibinfo{person}{Yi Sun}, \bibinfo{person}{Tony Worm}, {and}
  \bibinfo{person}{Michael Reale}.} \bibinfo{year}{2008}\natexlab{}.
\newblock \showarticletitle{A high-resolution 3D dynamic facial expression
  database}. In \bibinfo{booktitle}{\emph{2008 8th IEEE International
  Conference on Automatic Face Gesture Recognition}}. \bibinfo{pages}{1--6}.
\newblock
\urldef\tempurl%
\url{https://doi.org/10.1109/AFGR.2008.4813324}
\showDOI{\tempurl}


\bibitem[Yin et~al\mbox{.}(2006)]%
        {yin20063d}
\bibfield{author}{\bibinfo{person}{Lijun Yin}, \bibinfo{person}{Xiaozhou Wei},
  \bibinfo{person}{Yi Sun}, \bibinfo{person}{Jun Wang}, {and}
  \bibinfo{person}{Matthew~J Rosato}.} \bibinfo{year}{2006}\natexlab{}.
\newblock \showarticletitle{A 3D facial expression database for facial behavior
  research}. In \bibinfo{booktitle}{\emph{7th international conference on
  automatic face and gesture recognition (FGR06)}}. IEEE,
  \bibinfo{pages}{211--216}.
\newblock


\bibitem[Yoon et~al\mbox{.}(2019)]%
        {yoon2019self}
\bibfield{author}{\bibinfo{person}{Jae~Shin Yoon}, \bibinfo{person}{Takaaki
  Shiratori}, \bibinfo{person}{Shoou-I Yu}, {and} \bibinfo{person}{Hyun~Soo
  Park}.} \bibinfo{year}{2019}\natexlab{}.
\newblock \showarticletitle{Self-supervised adaptation of high-fidelity face
  models for monocular performance tracking}. In
  \bibinfo{booktitle}{\emph{Proceedings of the IEEE/CVF Conference on Computer
  Vision and Pattern Recognition}}. \bibinfo{pages}{4601--4609}.
\newblock


\bibitem[Zhang et~al\mbox{.}(2021)]%
        {zhang2021neural}
\bibfield{author}{\bibinfo{person}{Longwen Zhang}, \bibinfo{person}{Qixuan
  Zhang}, \bibinfo{person}{Minye Wu}, \bibinfo{person}{Jingyi Yu}, {and}
  \bibinfo{person}{Lan Xu}.} \bibinfo{year}{2021}\natexlab{}.
\newblock \showarticletitle{Neural Video Portrait Relighting in Real-Time via
  Consistency Modeling}. In \bibinfo{booktitle}{\emph{Proceedings of the
  IEEE/CVF International Conference on Computer Vision (ICCV)}}.
  \bibinfo{pages}{802--812}.
\newblock


\bibitem[Zhang et~al\mbox{.}(2014b)]%
        {zhang2014bp4d}
\bibfield{author}{\bibinfo{person}{Xing Zhang}, \bibinfo{person}{Lijun Yin},
  \bibinfo{person}{Jeffrey~F Cohn}, \bibinfo{person}{Shaun Canavan},
  \bibinfo{person}{Michael Reale}, \bibinfo{person}{Andy Horowitz},
  \bibinfo{person}{Peng Liu}, {and} \bibinfo{person}{Jeffrey~M Girard}.}
  \bibinfo{year}{2014}\natexlab{b}.
\newblock \showarticletitle{Bp4d-spontaneous: a high-resolution spontaneous 3d
  dynamic facial expression database}.
\newblock \bibinfo{journal}{\emph{Image and Vision Computing}}
  \bibinfo{volume}{32}, \bibinfo{number}{10} (\bibinfo{year}{2014}),
  \bibinfo{pages}{692--706}.
\newblock


\bibitem[Zhang et~al\mbox{.}(2014a)]%
        {zhang2014facial}
\bibfield{author}{\bibinfo{person}{Zhanpeng Zhang}, \bibinfo{person}{Ping Luo},
  \bibinfo{person}{Chen~Change Loy}, {and} \bibinfo{person}{Xiaoou Tang}.}
  \bibinfo{year}{2014}\natexlab{a}.
\newblock \showarticletitle{Facial landmark detection by deep multi-task
  learning}. In \bibinfo{booktitle}{\emph{European conference on computer
  vision}}. Springer, \bibinfo{pages}{94--108}.
\newblock


\bibitem[Zollh{\"o}fer et~al\mbox{.}(2018)]%
        {zollhofer2018state}
\bibfield{author}{\bibinfo{person}{Michael Zollh{\"o}fer},
  \bibinfo{person}{Justus Thies}, \bibinfo{person}{Pablo Garrido},
  \bibinfo{person}{Derek Bradley}, \bibinfo{person}{Thabo Beeler},
  \bibinfo{person}{Patrick P{\'e}rez}, \bibinfo{person}{Marc Stamminger},
  \bibinfo{person}{Matthias Nie{\ss}ner}, {and} \bibinfo{person}{Christian
  Theobalt}.} \bibinfo{year}{2018}\natexlab{}.
\newblock \showarticletitle{State of the art on monocular 3D face
  reconstruction, tracking, and applications}. In
  \bibinfo{booktitle}{\emph{Computer Graphics Forum}},
  Vol.~\bibinfo{volume}{37}. Wiley Online Library, \bibinfo{pages}{523--550}.
\newblock


\end{thebibliography}

\end{document}